\documentclass[11pt]{article}

\usepackage[letterpaper,margin=0.75in]{geometry}
\usepackage[T1]{fontenc}
\usepackage{lmodern}
\usepackage{microtype}
\usepackage{wrapfig}
\usepackage{booktabs}

\usepackage{amsmath,amssymb,amsthm,mathtools,bm}
\usepackage{thmtools,thm-restate}

\PassOptionsToPackage{numbers,compress}{natbib}
\usepackage{natbib}

\usepackage[table]{xcolor}
\definecolor{ourscolor}{RGB}{230,243,255}  
\definecolor{mydarkblue}{RGB}{0,80,160}
\usepackage{hyperref}
\hypersetup{colorlinks=true, linktocpage}

\usepackage[capitalize,noabbrev]{cleveref}

\hypersetup{
    colorlinks,
    linkcolor={red!50!black},
    citecolor={blue!80!black},
    urlcolor={blue!80!black}
}

\theoremstyle{plain}

\theoremstyle{definition}

\theoremstyle{remark}

\usepackage{graphicx}
\usepackage{float}
\usepackage{subcaption}
\usepackage{enumitem}

\usepackage[most]{tcolorbox}

\newtcolorbox{summarybox}{
  colframe=black!70,   
  boxrule=1.5pt,         
  arc=4pt,                   
  left=2pt,
  right=2pt,
  top=3pt,
  bottom=3pt,
  halign=center,
}

\title{
  \textbf{Embedding Geometry Shapes LLM Routing: \\ Tracing Failure Modes and Improving Generalization}
}

\author{
  Leonardo Torti$^{1,\ }$\footnote{Main contributors. Correspondence to \texttt{\{pierb,leotorti\}@mit.edu}.
  \\ \phantom{a} \!\!\! \ \ $^{1}$Massachusetts Institute of Technology.
  \\ \phantom{a} \!\!\! \ \ $^{2}$Texas A\&M University.} 
  \quad Emanuele Rimoldi$^{1,*}$ \quad Tomer Galanti$^{2}$ \quad Tomaso Poggio$^{1}$ \\[0.4em]
  Pierfrancesco Beneventano$^{1}$
}

\date{}

\begin{document}

\maketitle
                  
\vspace{-0.1cm}

\begin{abstract}
With the growing number of LLMs available, selecting which one to use for a prompt has become a key question. A large part of the routing literature relies on semantically embedding prompts into a vector space and then applying a lightweight classifier to the embedded representations. These methods perform robustly in-distribution but they struggle to generalize to unseen tasks. By studying the geometry of semantic embeddings, we trace a clear failure mode: features that determine the neighbors of unseen prompts are an unreliable proxy for LLM selection. We therefore replace the semantic embedding with the following map: we run a partial forward pass of Qwen2.5-7B-Instruct on the prompt and use the hidden state of its last token at a deep layer as the representation. When applying kNN and other simple classifiers to this new representation, we obtain a clear result in generalization: on average across cost penalties, their performance improves. Moreover, kNN and another simple classifier that we propose outperform every evaluated semantic-embedding baseline. Our two simple methods also beat a recent SoTA hidden-states based router on RouterBench.
\end{abstract}


\section{Introduction}
\label{sec:intro}

A large number of LLMs are now available and post-training your own open-weight model is becoming increasingly accessible. This raises questions about which \textbf{fundamental principles} should govern LLM selection, commonly called LLM routing, when addressing a specific task. We tackle the following problem:
\vspace{0.03cm}
\begin{center}
    \textit{Given an evaluation suite, construct a router that selects the optimal model\\ for the tasks seen at deployment time.}
    \vspace{0.01cm}
\end{center}
The evaluation suite records the reward (e.g., accuracy) and cost of each available model on each query in a given set. This set of queries is usually made of different task families (Section~\ref{sec:framework}). Using this data, the task is to train a classifier, called a router, that takes a prompt as input and selects which model to use.

We focus on compute-aware routing: methods that do not rely on reasoning. Most leading compute-aware approaches first map each prompt to a vector space using a semantic text encoder, then train a classifier on these representations to select a model.

We study their in-distribution (ID) performance, where training queries are representative of the deployment distribution, and out-of-distribution (OOD) performance, where that distribution changes.
In-distribution, popular compute-aware methods appear robust, and many achieve similar performance, with simple methods such as kNN often matching more complex approaches \citep{li2025rethinking,lu2026routingplateau}. Out of distribution, the differences between methods become more pronounced, and most routers struggle.
\begin{summarybox}
    \textbf{Our objective is to unveil:} \textbf{(i)} \textit{the mechanism underlying the failure modes of these methods};\\ \textbf{(ii)} \textit{what causes these more or less pronounced differences}; and \\ \textbf{(iii)} \textit{how to leverage those to improve the current methods}.
\end{summarybox}
\vspace{-0.4cm}
\clearpage


\begin{figure}[h!]
\centering
\includegraphics[width=\linewidth]{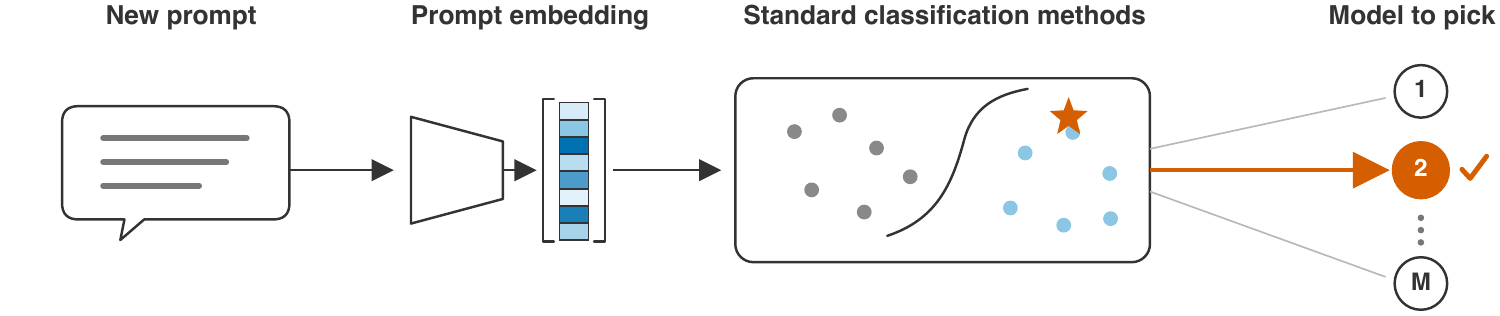}
\caption{Compute-aware routing: a prompt is mapped to a representation, on which a standard classifier selects the model to use.}
\label{fig:pipeline}
\end{figure}

Through the geometry of query representations, we ask whether features shaping locality also inform model selection. Semantic proximity can retrieve queries that provide unreliable guidance on unseen tasks. This motivates replacing semantic embeddings with frozen LLM hidden states while keeping routing rules fixed. The resulting gains let simple routers compete with substantially more elaborate pipelines, and our geometric analysis links them to changes in the features shaping locality. \newline
In summary, our contributions are:
\begin{itemize}[leftmargin=1.2em,itemsep=3pt,topsep=4pt,parsep=0pt]
\item \textbf{ID: explaining why simple routers remain competitive.} Our geometric analysis shows that semantic embeddings are strongly organized by family and topic. This organization explains why simple neighborhood methods and rules that select a single model for each family or finer group can match more complex routers. 

\item \textbf{OOD: tracing the failure mode to misleading neighborhoods.} We show that queries from an unseen family can concentrate near a seen family close in topic or prompt format, but whose model-performance information poorly guides selection for those queries; correspondingly, semantic kNN does no better than always selecting the best single model. With the semantic encoder fixed, we introduce Family Adaptation: a multinomial logistic classifier weights seen families per query, and the router combines their average per-model rewards and costs. Bypassing potentially misleading nearest-family neighborhoods, this simple rule outperforms all evaluated external routers using GTE on RouterBench and all except Ridge on SPROUT.

\item \textbf{Improving generalization through a change of representation.} Building on our analysis of semantic-embedding failure modes and prior work proposing routing methods with LLM hidden states, we replace semantic embeddings with a frozen LLM's last-token state from a single internal layer, keeping routing algorithms unchanged. This improves held-out-family routing without further adaptation: kNN-100 and Family Adaptation each outperform every evaluated external semantic-embedding baseline in mean utility across cost penalties on RouterBench and SPROUT. Beyond these popular and strong baselines, we compare our simple, interpretable pipelines with SharedTrunkNet \citep{varshney2026prefill}, a recent state-of-the-art router based on LLM hidden states. We use the same Qwen2.5-7B-Instruct encoder for all methods to isolate pipeline differences. Family Adaptation clearly outperforms SharedTrunkNet on RouterBench and trails it only slightly on SPROUT; even kNN-100 outperforms it on RouterBench and remains close behind on SPROUT.

\item \textbf{Understanding the geometric change.}
Keeping routing rules fixed lets us attribute the gains to the representation change. Compared with semantic embeddings, hidden-state embeddings much more often retrieve neighbors from training families sharing the unseen query's answer format. This suggests a possible explanation for semantic routers' strong ID performance: answer format is often shared within a family, so neighbors from that family share it even when other features shape locality. Holding out the query's family breaks this association, exposing whether those features retrieve examples useful for routing. Still, greater answer-format agreement does not fully explain the gains.
\end{itemize}

\section{Background and Notation}
\label{sec:framework}

\paragraph{The routing problem.}
Let $\mathcal{X}$ denote the space of prompts, which we also call queries, and let $\mathcal{M}=\{m_1,\ldots,m_M\}$ be a fixed pool of candidate models. We evaluate the routing problem on a benchmark $\mathcal{S}=\{x_i\}_{i=1}^{n}\subset\mathcal{X}$, which is a set of prompts. Routing benchmarks, including those studied here, are essentially compositions of several individual benchmarks, such as MMLU and GSM8K. We call each constituent benchmark a family and denote its queries by $\mathcal{S}_j$, so that $\mathcal{S}=\bigcup_{j=1}^{J}\mathcal{S}_j$, where $J$ is the number of families. For every query $x\in \mathcal{S}$, we have a reward $r(x,m)\in[0,1]$, measuring how well $m$ solves $x$, and the cost $c(x,m)\in\mathbb{R}_{\ge0}$ of the response in US dollars. Binary accuracy is the special case $r(x,m)\in\{0,1\}$. For a cost penalty $\lambda\ge0$, the utility of answering $x$ with $m$ is
\[
u_\lambda(x,m)=r(x,m)-\lambda c(x,m).
\]
A router is a map $\pi_\lambda:\mathcal{X}\rightarrow\mathcal{M}$ that, given a prompt, selects a model to answer it before observing any candidate's response. The routing problem is to construct, from the training data composed of query-model pairs and their measured utilities, a router whose selections maximize the utility on the prompts encountered at deployment. Routers that predict rewards and costs select
\[
\pi_\lambda(x) \in
\arg\max_{m\in\mathcal{M}}
\bigl[\hat r(x,m)-\lambda\hat c(x,m)\bigr],
\]
where $\hat r$ and $\hat c$ are estimates obtained from training data. For a held-out set of prompts $\mathcal{T}$, we evaluate the empirical utility
\[
\widehat U_\lambda(\pi_\lambda;\mathcal{T})
=
\frac{1}{|\mathcal{T}|}
\sum_{x\in\mathcal{T}}
u_\lambda\bigl(x,\pi_\lambda(x)\bigr),
\]
computed from the recorded reward and cost of each selected model, without generating new responses. We also report mean utility averaged equally across the 13 cost penalties.

\paragraph{Embeddings.}
A semantic embedding is a vector $\mathcal{F}(x)\in\mathbb{R}^{d}$ produced by a frozen text encoder $\mathcal{F}:\mathcal{X}\rightarrow\mathbb{R}^{d}$ trained to map texts with similar meaning to nearby vectors. The classification algorithm, e.g., $k$-nearest neighbors, logistic regression, an MLP, or clustering, then maps $\mathcal{F}(x)$ to a model in $\mathcal{M}$. Our geometric study uses \texttt{GTE-Qwen2-7B-instruct} \citep{alibaba2024gteqwen2,li2023gte} as the main semantic encoder; additional results for \texttt{all-MiniLM-L12-v2}~\citep{sentencetransformers_minilm_l12_v2} appear in Appendix~\ref{app:family-topic-geometry}. In Section~\ref{sec:encoder}, we replace $\mathcal{F}$ with the embedding map $\mathcal{H}:\mathcal{X}\to\mathbb{R}^{d}$, derived from \texttt{Qwen2.5-7B-Instruct} \citep{qwen2025technical} hidden states and defined there in detail. \newline
We will compare prompts through the cosine similarity of their representations, based on their normalized representation $z(x)=\mathcal{F}(x)/\|\mathcal{F}(x)\|_2$, and analogously for $\mathcal{H}$. The neighborhood $\mathcal{N}_k(x;\mathcal{R})$ of a prompt $x$ is the set of the $k$ most similar prompts in a reference pool $\mathcal{R}$, excluding $x$ itself. Routing uses training prompts as the reference pool for neighborhoods; geometric diagnostics specify their pools separately.

\paragraph{Datasets and evaluation protocols.}
We use two leading routing benchmarks. RouterBench \citep{hu2024routerbench} consists of 34,771 prompts from seven source datasets (MMLU, HellaSwag, GSM8K, ARC, WinoGrande, MBPP and MT-Bench), with recorded rewards and dollar costs for 11 models. SPROUT \citep{somerstep2025carrot} consists of 43,887 prompts from six source datasets (GPQA, MATH, MMLU-Pro, MuSR, OpenHermes and RAGBench), with rewards computed for 13 models; we compute its costs from the released token counts and fixed model prices. These datasets also support finer subdivisions of families, defined by topic labels or other native labels.  We will refer to these subdivisions as cells, or topics when the cell represents a topic. Families without finer labels form a single cell. Appendix~\ref{app:protocol} gives full descriptions, label provenance and preprocessing.
For in-distribution evaluation, we use five family-stratified 70/30 train-test splits and average the scores across splits. For OOD evaluation, we hold out each family in turn, train on the remaining families, and pool the held-out-family predictions across prompts. Compared methods share the same outer splits; router hyperparameters are fixed or selected using training data only, with method-specific validation procedures detailed in Appendix~\ref{app:protocol}. 

\section{On Semantic Embeddings}
\subsection{Related Work}
\begin{wrapfigure}{r}{0.55\linewidth}
\vspace{-2.5cm}
\centering
\includegraphics[width=\linewidth]{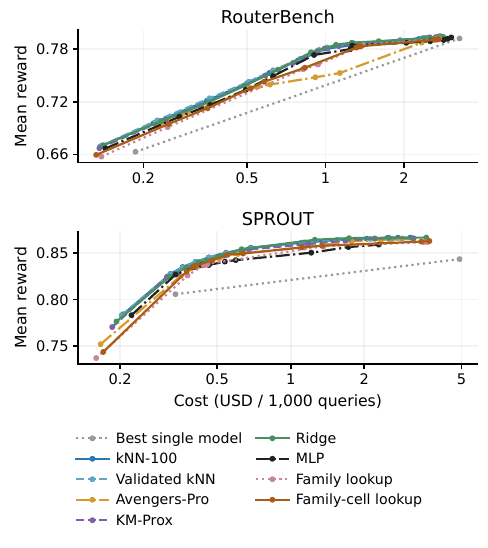}
\caption{In-distribution reward--cost frontiers with GTE. Neighborhood methods and group lookup remain competitive with more elaborate routers.}
\label{fig:iid-frontiers}
\vspace{-0.5cm}
\end{wrapfigure}
Simple routers can match more complex methods when routing in distribution.
\citet{li2025rethinking} demonstrates the competitiveness of kNN,
while \citet{lu2026routingplateau} show that many different routing
methods achieve nearly identical ID performance, see also \citep{li2026llmrouterbench}.
\citet{lee2026tasktype} further highlights the predictive value of
task type. These findings motivate our geometric analysis, which
identifies the features shaping semantic neighborhoods and explains
why simple routing rules remain competitive in distribution.
This understanding then provides the starting point for investigating
what fails when queries come from unseen families.

Generalization to unseen tasks presents additional challenges.
\citet{shnitzer2023routing} study routing to new tasks through
model-correctness prediction. RouteLLM~\citep{ong2024routellm}
learns from preference data and evaluates generalization to other
benchmarks and model pairs. ProxRouter~\citep{patel2025proxrouter}
improves robustness to outlier tasks by weighting model-performance
estimates according to proximity and uncertainty; its KM-Prox
variant is included in our comparisons. Our analysis examines
the geometric mechanism behind generalization failures.
While keeping the semantic embedding fixed, our diagnosis motivates our Family Adaptation method. Related aggregation strategies appear in
UniRoute~\citep{jitkrittum2025universal}, which combines cluster-level
model-performance estimates, and TRouter~\citep{liu2026trouter},
which combines task-specific reward and cost predictions using an MLP.
Family Adaptation is also inspired by distribution-weighted combinations in multiple-source adaptation \citep{mansour2008adaptation,hoffman2018adaptation}.

\subsection{In-Distribution Performance Matches Family-Cell Lookup}
\label{sec:iid-lookup}
kNN matches or outperforms many evaluated routers, consistent with \citet{lu2026routingplateau}.
Even family and family-cell lookup remain competitive. To route a query, family lookup selects the model with the highest mean training utility within that query's family; family-cell lookup applies the same rule within its cell. Both use the query's known group label. See Figure \ref{fig:iid-frontiers}.
We further test whether kNN improves utility when its choices differ from family-cell lookup, finding small mean gains on both benchmarks (Appendix~\ref{app:iid-lookup-disagreement}).

\subsection{How Geometry Shapes ID Routing}
\label{sec:iid-geometry}
GTE neighborhoods strongly preserve family identity, helping explain why neighborhood routing and group lookup achieve similar in-distribution performance (Section~\ref{sec:iid-lookup}). We report results for GTE-Qwen2-7B-Instruct on RouterBench and SPROUT. Appendix~\ref{app:family-topic-geometry} repeats the family and subject geometry tests with the semantic encoder all-MiniLM-L12-v2.
We see that 95.5\% of the 100 nearest training examples to a prompt share its family on RouterBench and 97.3\% on SPROUT. Within families, subjects (or in general cells) provide a
further level of organization (Figure~\ref{fig:geometry-hierarchy});
tests can be found in Appendix~\ref{app:family-topic-geometry}.

Many other query properties could shape locality. As one example, we examine difficulty, measured by the fraction of models achieving reward at least 0.5, with higher fractions indicating easier queries. Within each family-cell, neighbors have difficulty closer to that of the query than randomly selected examples do. The difference is small on RouterBench and larger on SPROUT (Appendix~\ref{app:difficulty-locality}).

\begin{figure*}[!htbp]
\centering
\includegraphics[width=\linewidth]{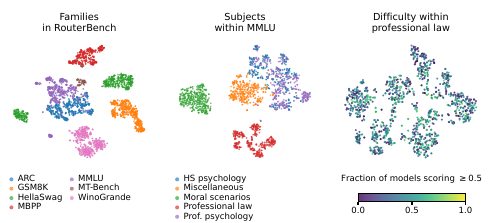}
\caption{GTE geometry on RouterBench: families, subjects within MMLU,
and difficulty within professional law. Each panel uses a separate
t-SNE projection.}
\label{fig:geometry-hierarchy}\vspace{-4.4pt}
\end{figure*}

\subsection{Held-Out Families Expose Misleading Semantic Neighborhoods}\label{sec:ood}

\begin{figure}[h]
\centering
\includegraphics[width=0.75\linewidth]{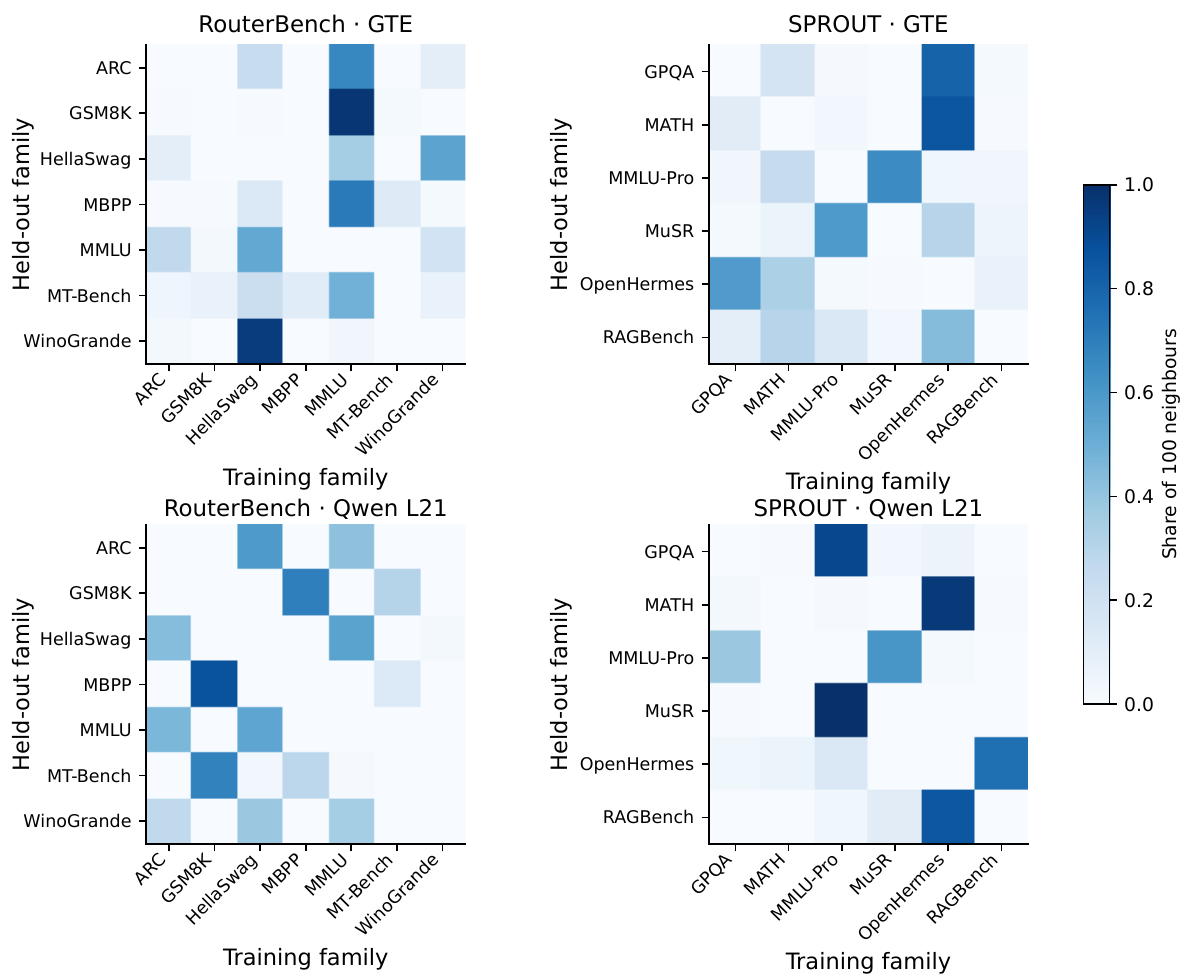}
\caption{Changing the representation changes which training families supply held-out queries' neighbors. Each matrix reports the family composition of the 100 nearest training queries under held-out-family evaluation.}
\label{fig:ood-family-composition}
\end{figure}

We saw that in-distribution routing can rely strongly on family structure, particularly for locality-based methods. We now examine what happens when the query's family is absent from training.
Removing all other members of a query's family redirects its neighborhood to essentially another single family. On average, a single training family supplies 78.4\% of each held-out query's 100 nearest neighbors on RouterBench and 84.1\% on SPROUT (Appendix~\ref{app:family-concentration}). Figure~\ref{fig:ood-family-composition} shows which training families are retrieved.

Concentration alone does not establish a routing failure.
We therefore test whether the families attracting neighbors
provide useful evidence for LLM selection.
For each held-out family and cost penalty, we predict model
utilities using neighbors from each source family separately.
We then ask whether the family that best predicts utility differences between candidate models is also the one supplying the most neighbors.
These families rarely coincide (Appendix~\ref{app:transfer}).

Consistent with this mismatch, GTE kNN-100 achieves mean utility across cost penalties of
0.6615 on RouterBench and 0.7529 on SPROUT, compared with
0.6618 and 0.7575 obtained by simply always choosing the best single model selected from
source training data separately for each penalty and outer split.
Together, these results show that GTE neighborhood structure
does not reliably identify useful model-selection evidence
when the query's family is absent from training.
The representation comparison in Section~\ref{sec:encoder} tests whether this failure is specific to semantic embeddings.

\subsection{A Simple Classifier: Family Adaptation}
On both benchmarks, Family Adaptation improves mean held-out-family utility over GTE kNN-100 while keeping the embedding fixed. Inspired by multiple-source domain adaptation \citep{mansour2008adaptation}, it combines performance information from several training families. We train a multinomial logistic classifier on the observed training-family labels and use its output probabilities $p(f\mid x)$ as weights over those families for an unseen query $x$. For each model, we combine the families' mean training utilities:
\[
\widehat u_\lambda(x,m)
=\sum_f p(f\mid x)\left(\bar r_{fm}-\lambda\bar c_{fm}\right).
\]
Here, $\bar r_{fm}$ and $\bar c_{fm}$ are model $m$'s mean reward and cost on training queries in family $f$. The router selects the model with the largest estimate. The classifier has balanced class weights and fixed regularization; the reward and cost averages use only training queries. Test family labels are not required.

This rule retains coarse performance information while allowing several families to contribute. Using GTE, it raises mean held-out utility over the penalty grid from 0.6615 to 0.7006 on RouterBench and from 0.7529 to 0.7603 on SPROUT relative to kNN-100 (Appendix~\ref{app:results}).

\section{Hidden-State Embeddings}\label{sec:encoder}
Let $x$ be a prompt. We now consider the embedding $\mathcal{H}:\mathcal{X}\to \mathbb{R}^{3584}$, where $\mathcal{H}(x)$ is the hidden state of the last input token at layer 21 of Qwen2.5-7B-Instruct. This representation can be computed by running the forward pass only through layer 21, without generating an answer. L21 was fixed following exploratory RouterBench analysis; Appendix~\ref{app:layer-sweep} shows results' robustness across layers. 
\subsection{Related Work}
\label{sec:hidden-states-related-work}

Recent work has proposed methods based on LLM hidden states for correctness prediction and routing.
\citet{morenocencerrado2025noanswer} predict  correctness from
question-only activations, while \citet{lugoloobi2026failures}
\begin{minipage}[t]{0.43\textwidth}
\vspace{0pt}
use pre-generation success predictions for model selection.
\citet{chen2026intent} use activation statistics to classify
prompt intent.
\citet{wu2026probedirichlet} introduce ProbeDirichlet for escalation
from a small model to a larger one using LLM hidden states.
\citet{varshney2026prefill} use a separate encoder's prefill
activations and an ensemble of neural predictors to estimate
correctness across candidate models.
\citet{pichlmeier2026funnel} introduce Funnel Pooling, a learned
aggregation of hidden states for LoRA-expert routing, trained on question--answer pairs.
For queries in different languages, they observe that standard pooling can yield clusters organized more by language and format than by topic, and propose a pooling that produces more topic-aligned
clusters. 
We build on this line of work from a different angle: rather than designing tailored predictors, we show that simple classifiers on hidden states can be competitive with more complex routers, and try to explain why.

\subsection{Changing the Embedding \\Improves Routing} \label{sec:embedding-improves-routing}

Replacing GTE with L21 raises mean held-out-family utility for every evaluated routing rule on both benchmarks except Avengers-Pro on RouterBench (Table~\ref{tab:ood-encoder-means}). We compare the representations while keeping the queries, candidate models, routing rules and evaluation splits fixed.

Family structure persists: the training family supplying the most neighbors to each held-out query accounts, on average, for 72.7\% of its 100 nearest neighbors on RouterBench and 92.0\% on SPROUT, compared with 78.4\% and 84.1\% under $\mathcal{F}$ (Appendix~\ref{app:family-concentration}).

We will see in the following paragraphs that the retrieved families are not only different but also a better proxy for routing on held-out families. A brief check in Appendix~\ref{app:ordering} also shows that L21 predicts pairwise model-utility orderings more accurately than GTE.
\vspace{0.1cm}
\end{minipage}\hfill%
\begin{minipage}[t]{0.55\textwidth}
\vspace{-0.3cm}
\centering
\captionsetup{type=figure}
\includegraphics[width=\linewidth,height=0.78\textheight,keepaspectratio]{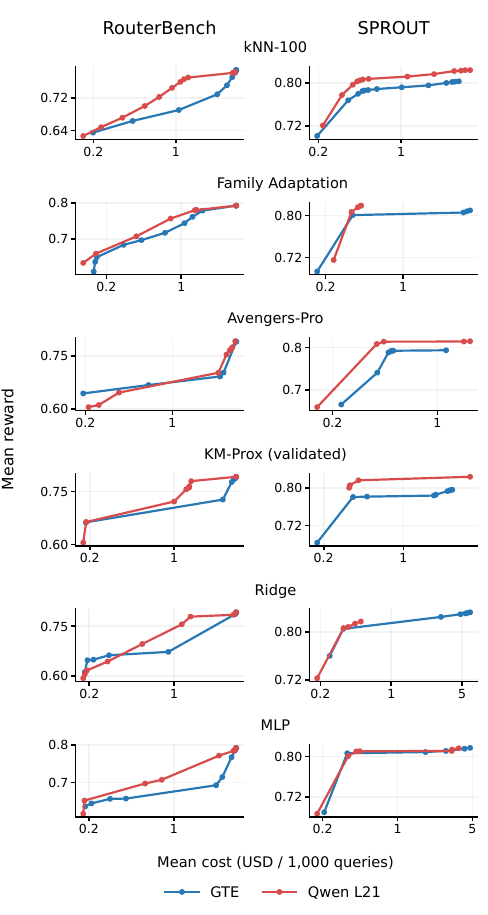}
\caption{Matched routing rules under held-out-family evaluation: GTE in blue and L21 in red. The representation change improves mean utility for most evaluated rules; Table~\ref{tab:ood-encoder-means} gives the averages and Appendix~\ref{app:results} reports every penalty.}
\label{fig:matched}
\end{minipage}

Figure~\ref{fig:matched} compares identical routing rules across the two representations, while Table~\ref{tab:ood-encoder-means} reports mean utility across all cost penalties. kNN-100 improves at nine of thirteen penalties on RouterBench and all thirteen on SPROUT. Its mean utility rises from 0.6615 to 0.7007 and from 0.7529 to 0.7731, respectively. Results by held-out family, with equal-query and equal-family averages, and paired confidence intervals, are reported in Appendix~\ref{app:routing-gain-by-family}.

To put this improvement in perspective, the GTE baselines (excluding our introduced Family Adaptation) span mean utilities from 0.6566 to 0.6796 on RouterBench and 0.7439 to 0.7674 on SPROUT across the thirteen cost penalties, a spread of 0.0230 and 0.0235, respectively. The gain of kNN with the embedding change is 0.0391 and 0.0202, respectively.

The improvement extends beyond kNN: mean utility increases for every evaluated routing rule on both benchmarks, except Avengers-Pro on RouterBench. This suggests that hidden states provide more exploitable patterns for linear methods such as ridge regression and nonlinear methods such as MLP, as well as more informative neighbors. We emphasize kNN because it is the simplest method we measured and the most geometrically interpretable.

\begin{table}[h!]
\setlength{\abovecaptionskip}{4pt}
\setlength{\belowcaptionskip}{6pt}
\centering
\small
\setlength{\tabcolsep}{5pt}
\caption{L21 raises mean utility for every evaluated routing rule on both benchmarks except Avengers-Pro on RouterBench. Values average all 13 cost penalties under held-out-family evaluation; bold marks the higher value within each encoder pair. Full results with 95\% confidence intervals are reported in Appendix~\ref{app:results}.}
\label{tab:ood-encoder-means}
\begin{tabular}{@{}lrr@{\hspace{1.8em}}rr@{}}
\toprule
& \multicolumn{2}{c}{\textbf{RouterBench}} & \multicolumn{2}{c}{\textbf{SPROUT}} \\
\cmidrule(lr){2-3}\cmidrule(lr){4-5}
Method & GTE & L21 & GTE & L21 \\
\midrule
Best single model & \multicolumn{2}{c}{0.6618} & \multicolumn{2}{c}{0.7575} \\
kNN-100 & 0.6615 & \textbf{0.7007} & 0.7529 & \textbf{0.7731} \\
Validated kNN & 0.6566 & \textbf{0.7034} & 0.7534 & \textbf{0.7739} \\
Family Adaptation & 0.7006 & \textbf{0.7157} & 0.7603 & \textbf{0.7767} \\
Avengers-Pro & \textbf{0.6662} & 0.6581 & 0.7439 & \textbf{0.7714} \\
KM-Prox (validated) & 0.6764 & \textbf{0.6984} & 0.7483 & \textbf{0.7764} \\
Ridge & 0.6796 & \textbf{0.6902} & 0.7674 & \textbf{0.7712} \\
MLP & 0.6755 & \textbf{0.6918} & 0.7564 & \textbf{0.7667} \\
\bottomrule
\end{tabular}
\end{table}

The L21 Family Adaptation achieves the highest observed mean utility among the methods in Table~\ref{tab:ood-encoder-means} on both benchmarks. L21 kNN-100 also exceeds every evaluated external GTE baseline in mean utility on both benchmarks (Table~\ref{tab:ood-mean}).

\begingroup
\clubpenalties 4 10000 10000 10000 0
\widowpenalty=10000
Finally, we compare L21 kNN-100 and L21 Family Adaptation with the SharedTrunkNet router of \citet{varshney2026prefill}, using a common frozen Qwen2.5-7B-Instruct encoder. We retain its architecture and its own feature preprocessing and layer selection, while adapting feature selection and network training to our utility objective. We follow the validation procedure as closely as possible (Appendix~\ref{app:sharedtrunknet}).\par
\endgroup

\begin{table}[H]
\setlength{\abovecaptionskip}{4pt}
\setlength{\belowcaptionskip}{6pt}
\centering
\small
\setlength{\tabcolsep}{4pt}
\caption{Common-encoder comparison with utility-adapted SharedTrunkNet. The simple L21 routers have higher mean utility on RouterBench; SharedTrunkNet has higher mean utility on SPROUT. Values average all 13 cost penalties under held-out-family evaluation; bold marks the highest mean. Full results with 95\% confidence intervals are reported in Appendix~\ref{app:sharedtrunknet-results}.}
\label{tab:nvidia-utility-comparison}
\begin{tabular}{@{}lrr@{}}
\toprule
Method & RouterBench & SPROUT \\
\midrule
L21 kNN-100 & 0.7007 & 0.7731 \\
L21 Family Adaptation & \textbf{0.7157} & 0.7767 \\
SharedTrunkNet & 0.6949 & \textbf{0.7829} \\
\bottomrule
\end{tabular}
\end{table}

Our methods achieve these results with minimal training: kNN-100 requires no fitted prediction model, while Family Adaptation fits only a linear classifier.

Furthermore, we note that most gains from the methods above come for $\lambda>0$. This hints at the possibility, which we conjecture, that the gains come from a better representation of cost by the embedding space. We will address this later in Section~\ref{sec:answer-format-utility}.

\subsection{What Is the Difference Between Hidden-State and Semantic Embeddings?}
We now investigate which possible features could be responsible for the generalization improvement achieved with hidden-state embeddings.
Here, answer format denotes whether a task is multiple-choice or free-form.
When retrieving neighbors for a query from a held-out family, the share of neighbors matching the query's answer format increases: from 76.4\% to 99.9\% on RouterBench and from 58.9\% to 90.2\% on SPROUT (Appendix~\ref{app:answer-format}).  It is also consistent with the objective of the two encoders: GTE is fine-tuned to produce embeddings for text similarity and retrieval, whereas Qwen2.5-Instruct is trained to generate responses to instructions \citep{alibaba2024gteqwen2,qwen2025technical}.
This finding also exposes a potential cause for why kNN and other methods work well in-distribution with semantic embeddings: within the same family, the answer format does not change in most benchmarks. Thus, family is a good proxy for answer format. When the query's family is held out, the failure mode appears: semantic proximity no longer reliably retrieves useful evidence for model selection.

Restricting GTE retrieval to queries sharing the test query's answer format recovers most of L21's mean utility gain on RouterBench, but much less on SPROUT. We predict reward and cost from the 100 nearest eligible training queries, with the test query's family excluded. This control changes both format agreement and the available source-family composition; it does not isolate answer format as the sole cause.

\begin{table}[H]
\setlength{\abovecaptionskip}{4pt}
\setlength{\belowcaptionskip}{6pt}
\centering
\small
\setlength{\tabcolsep}{4pt}
\caption{Restricting GTE retrieval by answer format recovers most of L21's mean utility gain on RouterBench and less on SPROUT. The control also changes source-family composition. Values average held-out-family kNN-100 utility across 13 cost penalties; full results and 95\% intervals appear in Appendix~\ref{app:answer-format-control}.}
\label{tab:answer-format-control}
\begin{tabular}{@{}lrr@{}}
\toprule
Embedding / retrieval & RouterBench & SPROUT \\
\midrule
GTE & 0.6615 & 0.7529 \\
GTE, same answer format & 0.6974 & 0.7584 \\
L21 & \textbf{0.7007} & \textbf{0.7731} \\
\bottomrule
\end{tabular}
\end{table}

L21 therefore provides useful routing information not recovered by this format-restricted GTE control. The additional query properties associated with the remaining gap remain to be identified.

\subsection{Reward and Cost Contributions Differ Across Benchmarks}
\label{sec:answer-format-utility}

\noindent
\begin{minipage}[t]{0.54\linewidth}
\vspace{0pt}
We now examine whether L21's routing gains come from changes in reward estimates, cost estimates, or both. Prior work shows that prefill activations contain information predictive of answer correctness \citep{morenocencerrado2025noanswer,varshney2026prefill}.
For each held-out query and candidate model, we predict reward and cost separately by averaging the recorded outcomes of its 100 nearest training queries.

The prediction gains differ across benchmarks: L21 improves reward prediction on both, while cost prediction improves on RouterBench and slightly worsens on SPROUT (Table~\ref{tab:reward-cost-prediction}).

To separate these contributions, we replace GTE's reward or cost estimates with L21's in kNN-100 and compare mean utility across the thirteen cost penalties (Table~\ref{tab:answer-format-utility}). Prediction error and routing utility need not move together: on SPROUT, L21 cost estimates have slightly higher mean squared error but yield a small utility gain when paired with GTE reward estimates.
\end{minipage}\hfill
\begin{minipage}[t]{0.42\linewidth}
\vspace{-0.2cm}
\begingroup
\captionsetup{type=table}
\setlength{\abovecaptionskip}{4pt}
\setlength{\belowcaptionskip}{6pt}
\centering
\small
\setlength{\tabcolsep}{5pt}
\caption{Reduction in mean squared error from GTE to L21 (\%). Negative values indicate higher error. Full results and 95\% intervals appear in Appendix~\ref{app:prediction-errors}.}
\label{tab:reward-cost-prediction}
\begin{tabular}{@{}lrr@{}}
\toprule
Prediction & RouterBench & SPROUT \\
\midrule
Reward & 13.6 & 22.2 \\
Cost & 59.0 & $-2.6$ \\
\bottomrule
\end{tabular}
\par\endgroup
\par\vspace{13pt}
\begingroup
\captionsetup{type=table}
\setlength{\abovecaptionskip}{4pt}
\setlength{\belowcaptionskip}{6pt}
\centering
\small
\setlength{\tabcolsep}{4pt}
\caption{Changing cost estimates yields the larger utility gain on RouterBench; changing reward estimates yields the larger gain on SPROUT. Full results and 95\% intervals appear in Appendix~\ref{app:combined-estimates}.}
\label{tab:answer-format-utility}
\begin{tabular}{@{}llrr@{}}
\toprule
Reward & Cost & RouterBench & SPROUT \\
\midrule
GTE & GTE & 0.6615 & 0.7529 \\
L21 & GTE & 0.6754 & 0.7709 \\
GTE & L21 & 0.6965 & 0.7560 \\
L21 & L21 & \textbf{0.7007} & \textbf{0.7731} \\
\bottomrule
\end{tabular}
\par\endgroup
\end{minipage}
\par\medskip

\section{Conclusion}
We study how the geometry of prompt representations shapes compute-aware LLM routing. On RouterBench and SPROUT, family and topic structure help explain why simple routers remain competitive in distribution. Holding out a family exposes a failure mode: semantic proximity can retrieve examples whose model-performance information transfers poorly.

These findings give a practical response to the introductory question of constructing a router from an evaluation suite. Family Adaptation combines performance evidence across training families, while replacing semantic embeddings with frozen LLM hidden states improves mean held-out-family utility for most evaluated routing rules. This improves routing utility on the evaluated held-out families without requiring a more elaborate routing rule.

We thus identify a concrete failure mode, use it to motivate effective routing changes, and partially explain the differences between representations. The answer-format control recovers part of the representation gain, while reward and cost controls reveal benchmark-dependent sources of improvement. A complete explanation of the gains and a general characterization of optimal routing remain open.

\section*{Limitations}
At $\lambda=0$, no evaluated router improves on the best-single-model baseline in our out-of-distribution comparison. Beating this baseline requires identifying queries on which the best model fails but models with lower average performance succeed. Since LLMs are stochastic response generators, cases where the best model fails but another succeeds may depend on the particular responses sampled. Collecting multiple responses for each model--query pair would help assess whether these routing opportunities are consistent across generations. This limitation is due to compute/cost constraints, and it is shared by the literature.

Our routing comparison also uses a single pair of semantic and hidden-state encoders, and the observed gains are demonstrated for the encoders considered. Whether these results extend to other encoder sizes remains unclear.

Finally, the format-restricted GTE control leaves part of L21's gain unexplained, especially on SPROUT. Identifying the additional query properties associated with these gains remains future work.

\section*{AI Use Statement}
We used generative AI tools to assist with implementing and debugging
methods, preparing and checking datasets, designing and refining
experiments, refining hypothesis, and analyzing results. We also used these tools for literature search, preparing
figures and tables, drafting and editing text, and formatting
references. The authors take responsibility for the final content,
methods, and conclusions.

\section*{Acknowledgment}
This work was supported by the Center for Brains, Minds and Machines (CBMM, NSF STC award CCF-1231216); and the McGovern Institute for Brain Research at MIT.

\newpage
\bibliography{references}
\bibliographystyle{unsrtnat}

\clearpage
\appendix
\pdfbookmark[1]{Guide to the Appendices}{appendix-reading-guide}
\section*{Guide to the Appendices}
Evaluation and preprocessing are documented in \hyperref[app:protocol]{Protocol Details}; uncertainty estimates are specified under \hyperref[app:routing-ci]{Confidence Intervals}.
\begin{itemize}[leftmargin=1.2em,topsep=0.25em]
\item \textbf{ID competitiveness:}
\hyperref[app:family-topic-geometry]{family and subject structure},
\hyperref[app:iid-lookup-disagreement]{decisions beyond group lookup}, and
\hyperref[app:lookup-deployability]{lookup with predicted group labels}.
\item \textbf{OOD failure and Family Adaptation:}
\hyperref[app:family-concentration]{neighborhood concentration},
\hyperref[app:transfer]{model-performance transfer}, and
\hyperref[app:results]{full routing results}.
\item \textbf{Representation gains:}
\hyperref[app:results]{results across all penalties},
\hyperref[app:sharedtrunknet]{the SharedTrunkNet protocol}, and
\hyperref[app:layer-sweep]{layer choice and robustness}.
\item \textbf{Sources of improvement:}
\hyperref[app:answer-format]{answer-format agreement},
followed by the format-restriction and reward--cost controls.
\end{itemize}
\newsavebox{\resultoriginal}
\newsavebox{\resultstackprobe}
\newsavebox{\resultvalueprobe}
\newcommand{\tableminus}{\smash{$-$}\vphantom{-}}
\providecommand{\resultci}[3]{%
  \begingroup
  \sbox{\resultoriginal}{\shortstack[r]{#1\\[-1pt]{\scriptsize[#2,\,#3]}}}%
  \raisebox{0pt}[\ht\resultoriginal][\dp\resultoriginal]{%
    \shortstack[r]{\raisebox{0pt}[\height][0pt]{#1}\\[-1pt]{\scriptsize[#2,\,#3]}}}%
  \endgroup
}
\newcommand{\resultlabel}[1]{%
  \begingroup
  \sbox{\resultstackprobe}{\shortstack[r]{0\\[-1pt]{\scriptsize[0,\,0]}}}%
  \sbox{\resultvalueprobe}{0}%
  \raisebox{\dimexpr\ht\resultstackprobe-\ht\resultvalueprobe\relax}[\height][\depth]{#1}%
  \endgroup
}
\newcommand{\CIDeployability}{%
\begin{table}[H]
\setlength{\abovecaptionskip}{4pt}
\setlength{\belowcaptionskip}{6pt}
\centering\small
\setlength{\tabcolsep}{5pt}
\caption{Lookup with predicted labels, with 95\% intervals. Utilities average five splits and thirteen penalties; $\Delta$ is predicted-label minus known-label utility.}
\label{tab:lookup-deployability}
\begin{tabular}{@{}llrrrr@{}}
\toprule
Benchmark & Label & Accuracy & \shortstack{Known-label\\utility} & \shortstack{Predicted-label\\utility} & $\Delta$ \\[2pt]
\midrule
\resultlabel{RouterBench} & \resultlabel{Family} & \resultci{99.97\%}{99.95}{99.99} & \resultci{0.72124}{0.71720}{0.72523} & \resultci{0.72117}{0.71714}{0.72520} & \resultci{\tableminus 0.00007}{\tableminus 0.00018}{$+$0.00001} \\[2pt]
\resultlabel{RouterBench} & \resultlabel{Cell} & \resultci{93.85\%}{93.56}{94.14} & \resultci{0.72288}{0.71868}{0.72713} & \resultci{0.72232}{0.71819}{0.72646} & \resultci{\tableminus 0.00056}{\tableminus 0.00095}{\tableminus 0.00020} \\[2pt]
\resultlabel{SPROUT} & \resultlabel{Family} & \resultci{99.91\%}{99.88}{99.94} & \resultci{0.80966}{0.80609}{0.81320} & \resultci{0.80955}{0.80597}{0.81312} & \resultci{\tableminus 0.00012}{\tableminus 0.00026}{0.00000} \\[2pt]
\resultlabel{SPROUT} & \resultlabel{Cell} & \resultci{91.68\%}{91.38}{91.96} & \resultci{0.81225}{0.80881}{0.81584} & \resultci{0.81182}{0.80843}{0.81533} & \resultci{\tableminus 0.00042}{\tableminus 0.00100}{$+$0.00020} \\[2pt]
\bottomrule
\end{tabular}
\end{table}
}

\newcommand{\CIOodMean}{%
\begin{table}[H]
\setlength{\abovecaptionskip}{4pt}
\setlength{\belowcaptionskip}{6pt}
\centering\small
\setlength{\tabcolsep}{8pt}
\caption{Mean held-out utility across thirteen cost penalties, with 95\% intervals. Bold marks the highest mean among the listed methods.}
\label{tab:ood-mean}
\begin{tabular}{@{}lrr@{}}
\toprule
Method & RouterBench & SPROUT \\[2pt]
\midrule
\multicolumn{3}{@{}l}{\textit{GTE embeddings}} \\
\resultlabel{Best single model} & \resultci{0.6618}{0.6586}{0.6652} & \resultci{0.7575}{0.7544}{0.7604} \\[2pt]
\resultlabel{kNN-100} & \resultci{0.6615}{0.6583}{0.6651} & \resultci{0.7529}{0.7498}{0.7559} \\[2pt]
\resultlabel{Validated kNN} & \resultci{0.6566}{0.6534}{0.6601} & \resultci{0.7534}{0.7504}{0.7564} \\[2pt]
\resultlabel{Avengers-Pro} & \resultci{0.6662}{0.6631}{0.6698} & \resultci{0.7439}{0.7408}{0.7470} \\[2pt]
\resultlabel{KM-Prox (fixed)} & \resultci{0.6764}{0.6732}{0.6797} & \resultci{0.7460}{0.7429}{0.7491} \\[2pt]
\resultlabel{KM-Prox (validated)} & \resultci{0.6764}{0.6732}{0.6798} & \resultci{0.7483}{0.7453}{0.7513} \\[2pt]
\resultlabel{Ridge} & \resultci{0.6796}{0.6765}{0.6829} & \resultci{0.7674}{0.7645}{0.7703} \\[2pt]
\resultlabel{MLP} & \resultci{0.6755}{0.6722}{0.6789} & \resultci{0.7564}{0.7536}{0.7592} \\[2pt]
\resultlabel{Family Adaptation} & \resultci{0.7006}{0.6973}{0.7041} & \resultci{0.7603}{0.7573}{0.7632} \\[2pt]
\midrule
\multicolumn{3}{@{}l}{\textit{Qwen L21 embeddings}} \\
\resultlabel{kNN-100} & \resultci{0.7007}{0.6972}{0.7043} & \resultci{0.7731}{0.7701}{0.7760} \\[2pt]
\resultlabel{Family Adaptation} & \resultci{\textbf{0.7157}}{0.7125}{0.7191} & \resultci{\textbf{0.7767}}{0.7737}{0.7797} \\[2pt]
\bottomrule
\end{tabular}
\end{table}
}

\newcommand{\CINvidiaMean}{%
\begin{table}[H]
\setlength{\abovecaptionskip}{4pt}
\setlength{\belowcaptionskip}{6pt}
\centering\small
\setlength{\tabcolsep}{8pt}
\caption{Mean held-out utility across thirteen penalties, with 95\% intervals, for the comparison in the main text.}
\label{tab:nvidia-mean-ci}
\begin{tabular}{@{}lrr@{}}
\toprule
Method & RouterBench & SPROUT \\[2pt]
\midrule
\resultlabel{L21 kNN-100} & \resultci{0.7007}{0.6972}{0.7043} & \resultci{0.7731}{0.7701}{0.7760} \\[2pt]
\resultlabel{L21 Family Adaptation} & \resultci{0.7157}{0.7125}{0.7191} & \resultci{0.7767}{0.7737}{0.7797} \\[2pt]
\resultlabel{SharedTrunkNet} & \resultci{0.6949}{0.6918}{0.6983} & \resultci{0.7829}{0.7800}{0.7857} \\[2pt]
\bottomrule
\end{tabular}
\end{table}
}

\newcommand{\CIDisagreementSummary}{%
\begin{table}[H]
\setlength{\abovecaptionskip}{4pt}
\setlength{\belowcaptionskip}{6pt}
\centering\small
\setlength{\tabcolsep}{7pt}
\caption{GTE kNN-100 compared with group lookup across penalties, with 95\% paired intervals. Positive gains favor kNN.}
\label{tab:iid-disagreement-summary}
\begin{tabular}{@{}llrrr@{}}
\toprule
Dataset & Lookup & \shortstack{Different\\choices} & \shortstack{Gain on\\disagreements} & Overall gain \\[2pt]
\midrule
\resultlabel{RouterBench} & \resultlabel{Family} & \resultci{30.04\%}{29.80}{30.29} & \resultci{$+$0.0233}{$+$0.0178}{$+$0.0290} & \resultci{$+$0.0070}{$+$0.0054}{$+$0.0087} \\[2pt]
\resultlabel{RouterBench} & \resultlabel{Family-cell} & \resultci{28.78\%}{28.54}{29.03} & \resultci{$+$0.0186}{$+$0.0131}{$+$0.0245} & \resultci{$+$0.0054}{$+$0.0038}{$+$0.0071} \\[2pt]
\resultlabel{SPROUT} & \resultlabel{Family} & \resultci{45.50\%}{45.17}{45.79} & \resultci{$+$0.0199}{$+$0.0166}{$+$0.0233} & \resultci{$+$0.0090}{$+$0.0076}{$+$0.0106} \\[2pt]
\resultlabel{SPROUT} & \resultlabel{Family-cell} & \resultci{47.75\%}{47.44}{48.07} & \resultci{$+$0.0135}{$+$0.0101}{$+$0.0170} & \resultci{$+$0.0065}{$+$0.0048}{$+$0.0081} \\[2pt]
\bottomrule
\end{tabular}
\end{table}
}

\newcommand{\CIDisagreementPenalties}{%
\begin{table}[H]
\setlength{\abovecaptionskip}{4pt}
\setlength{\belowcaptionskip}{6pt}
\centering\small
\setlength{\tabcolsep}{9pt}
\caption{kNN-100 utility gains at every penalty, with 95\% paired intervals.}
\label{tab:iid-disagreement-penalties}
\begin{tabular}{@{}lrrrr@{}}
\toprule
& \shortstack{Against family lookup\\Overall gain} & \multicolumn{3}{c}{Against family-cell lookup} \\
\cmidrule(lr){3-5}
$\lambda$ &  & \shortstack{Different\\choices} & \shortstack{Gain on\\disagreements} & Overall gain \\[2pt]
\midrule
\multicolumn{5}{l}{\textbf{RouterBench}} \\
\resultlabel{0} & \resultci{0.0000}{\tableminus 0.0018}{$+$0.0019} & \resultci{23.55\%}{23.06}{24.03} & \resultci{$+$0.0038}{\tableminus 0.0039}{$+$0.0120} & \resultci{$+$0.0009}{\tableminus 0.0009}{$+$0.0029} \\[2pt]
\resultlabel{1} & \resultci{$+$0.0004}{\tableminus 0.0015}{$+$0.0023} & \resultci{24.92\%}{24.42}{25.42} & \resultci{$+$0.0057}{\tableminus 0.0018}{$+$0.0136} & \resultci{$+$0.0014}{\tableminus 0.0005}{$+$0.0034} \\[2pt]
\resultlabel{2} & \resultci{$+$0.0004}{\tableminus 0.0015}{$+$0.0022} & \resultci{25.51\%}{25.03}{25.99} & \resultci{$+$0.0062}{\tableminus 0.0012}{$+$0.0140} & \resultci{$+$0.0016}{\tableminus 0.0003}{$+$0.0036} \\[2pt]
\resultlabel{5} & \resultci{$+$0.0013}{\tableminus 0.0008}{$+$0.0033} & \resultci{29.27\%}{28.77}{29.79} & \resultci{$+$0.0078}{$+$0.0007}{$+$0.0149} & \resultci{$+$0.0023}{$+$0.0002}{$+$0.0043} \\[2pt]
\resultlabel{10} & \resultci{$+$0.0015}{\tableminus 0.0007}{$+$0.0038} & \resultci{22.00\%}{21.54}{22.47} & \resultci{$+$0.0115}{$+$0.0013}{$+$0.0222} & \resultci{$+$0.0025}{$+$0.0003}{$+$0.0049} \\[2pt]
\resultlabel{20} & \resultci{$+$0.0036}{$+$0.0013}{$+$0.0060} & \resultci{21.08\%}{20.62}{21.53} & \resultci{$+$0.0202}{$+$0.0095}{$+$0.0316} & \resultci{$+$0.0043}{$+$0.0020}{$+$0.0067} \\[2pt]
\resultlabel{30} & \resultci{$+$0.0061}{$+$0.0035}{$+$0.0088} & \resultci{24.78\%}{24.30}{25.25} & \resultci{$+$0.0263}{$+$0.0158}{$+$0.0365} & \resultci{$+$0.0065}{$+$0.0039}{$+$0.0091} \\[2pt]
\resultlabel{50} & \resultci{$+$0.0122}{$+$0.0098}{$+$0.0149} & \resultci{25.38\%}{24.96}{25.84} & \resultci{$+$0.0424}{$+$0.0330}{$+$0.0521} & \resultci{$+$0.0108}{$+$0.0083}{$+$0.0132} \\[2pt]
\resultlabel{75} & \resultci{$+$0.0114}{$+$0.0082}{$+$0.0146} & \resultci{27.53\%}{27.04}{28.03} & \resultci{$+$0.0289}{$+$0.0186}{$+$0.0392} & \resultci{$+$0.0080}{$+$0.0051}{$+$0.0108} \\[2pt]
\resultlabel{100} & \resultci{$+$0.0158}{$+$0.0122}{$+$0.0191} & \resultci{33.10\%}{32.57}{33.63} & \resultci{$+$0.0216}{$+$0.0125}{$+$0.0310} & \resultci{$+$0.0071}{$+$0.0041}{$+$0.0103} \\[2pt]
\resultlabel{150} & \resultci{$+$0.0148}{$+$0.0109}{$+$0.0186} & \resultci{39.11\%}{38.56}{39.62} & \resultci{$+$0.0226}{$+$0.0139}{$+$0.0316} & \resultci{$+$0.0088}{$+$0.0055}{$+$0.0124} \\[2pt]
\resultlabel{250} & \resultci{$+$0.0116}{$+$0.0081}{$+$0.0152} & \resultci{43.54\%}{42.98}{44.07} & \resultci{$+$0.0196}{$+$0.0121}{$+$0.0274} & \resultci{$+$0.0085}{$+$0.0052}{$+$0.0120} \\[2pt]
\resultlabel{500} & \resultci{$+$0.0119}{$+$0.0085}{$+$0.0154} & \resultci{34.39\%}{33.83}{34.92} & \resultci{$+$0.0198}{$+$0.0108}{$+$0.0287} & \resultci{$+$0.0068}{$+$0.0037}{$+$0.0099} \\[2pt]
\multicolumn{5}{l}{\textbf{SPROUT}} \\
\resultlabel{0} & \resultci{$+$0.0051}{$+$0.0027}{$+$0.0073} & \resultci{55.37\%}{54.88}{55.85} & \resultci{$+$0.0070}{$+$0.0026}{$+$0.0110} & \resultci{$+$0.0039}{$+$0.0015}{$+$0.0060} \\[2pt]
\resultlabel{0.5} & \resultci{$+$0.0054}{$+$0.0032}{$+$0.0077} & \resultci{57.35\%}{56.86}{57.83} & \resultci{$+$0.0078}{$+$0.0037}{$+$0.0115} & \resultci{$+$0.0045}{$+$0.0021}{$+$0.0066} \\[2pt]
\resultlabel{1} & \resultci{$+$0.0055}{$+$0.0032}{$+$0.0079} & \resultci{58.86\%}{58.38}{59.34} & \resultci{$+$0.0077}{$+$0.0036}{$+$0.0114} & \resultci{$+$0.0045}{$+$0.0021}{$+$0.0067} \\[2pt]
\resultlabel{2} & \resultci{$+$0.0066}{$+$0.0043}{$+$0.0089} & \resultci{58.08\%}{57.58}{58.56} & \resultci{$+$0.0102}{$+$0.0063}{$+$0.0142} & \resultci{$+$0.0059}{$+$0.0037}{$+$0.0082} \\[2pt]
\resultlabel{5} & \resultci{$+$0.0057}{$+$0.0031}{$+$0.0082} & \resultci{38.24\%}{37.80}{38.70} & \resultci{$+$0.0132}{$+$0.0071}{$+$0.0201} & \resultci{$+$0.0051}{$+$0.0027}{$+$0.0077} \\[2pt]
\resultlabel{10} & \resultci{$+$0.0064}{$+$0.0038}{$+$0.0089} & \resultci{35.35\%}{34.88}{35.81} & \resultci{$+$0.0143}{$+$0.0076}{$+$0.0210} & \resultci{$+$0.0051}{$+$0.0027}{$+$0.0075} \\[2pt]
\resultlabel{25} & \resultci{$+$0.0116}{$+$0.0093}{$+$0.0142} & \resultci{33.62\%}{33.17}{34.12} & \resultci{$+$0.0148}{$+$0.0088}{$+$0.0211} & \resultci{$+$0.0050}{$+$0.0029}{$+$0.0072} \\[2pt]
\resultlabel{50} & \resultci{$+$0.0079}{$+$0.0057}{$+$0.0102} & \resultci{35.87\%}{35.36}{36.35} & \resultci{$+$0.0072}{$+$0.0011}{$+$0.0135} & \resultci{$+$0.0026}{$+$0.0004}{$+$0.0049} \\[2pt]
\resultlabel{75} & \resultci{$+$0.0068}{$+$0.0047}{$+$0.0091} & \resultci{40.91\%}{40.44}{41.41} & \resultci{$+$0.0105}{$+$0.0053}{$+$0.0164} & \resultci{$+$0.0043}{$+$0.0022}{$+$0.0067} \\[2pt]
\resultlabel{100} & \resultci{$+$0.0070}{$+$0.0049}{$+$0.0093} & \resultci{45.56\%}{45.11}{46.07} & \resultci{$+$0.0092}{$+$0.0042}{$+$0.0143} & \resultci{$+$0.0042}{$+$0.0019}{$+$0.0065} \\[2pt]
\resultlabel{150} & \resultci{$+$0.0100}{$+$0.0077}{$+$0.0124} & \resultci{50.74\%}{50.28}{51.21} & \resultci{$+$0.0111}{$+$0.0066}{$+$0.0157} & \resultci{$+$0.0056}{$+$0.0033}{$+$0.0079} \\[2pt]
\resultlabel{250} & \resultci{$+$0.0157}{$+$0.0135}{$+$0.0181} & \resultci{53.11\%}{52.62}{53.58} & \resultci{$+$0.0200}{$+$0.0155}{$+$0.0246} & \resultci{$+$0.0106}{$+$0.0082}{$+$0.0130} \\[2pt]
\resultlabel{500} & \resultci{$+$0.0238}{$+$0.0205}{$+$0.0272} & \resultci{57.74\%}{57.27}{58.23} & \resultci{$+$0.0395}{$+$0.0343}{$+$0.0448} & \resultci{$+$0.0228}{$+$0.0198}{$+$0.0258} \\[2pt]
\bottomrule
\end{tabular}
\end{table}
}

\section{Protocol Details}\label{app:protocol}
\subsection{Data and evaluation protocol}
\paragraph{Data.}
RouterBench combines MMLU~\citep{hendrycks2020mmlu}, HellaSwag~\citep{zellers2019hellaswag}, GSM8K~\citep{cobbe2021gsm8k}, ARC~\citep{clark2018arc}, WinoGrande~\citep{sakaguchi2019winogrande}, MBPP~\citep{austin2021mbpp}, and MT-Bench~\citep{zheng2023mtbench}. SPROUT combines GPQA~\citep{rein2023gpqa}, MATH~\citep{hendrycks2021math}, MMLU-Pro~\citep{wang2024mmlupro}, MuSR~\citep{sprague2023musr}, OpenHermes~\citep{teknium2023openhermes25}, and RAGBench~\citep{friel2024ragbench}. We use the RouterBench zero-shot release, specifically \nolinkurl{routerbench_0shot.pkl}, and SPROUT at fixed dataset revisions.\footnote{Dataset snapshots: \href{https://huggingface.co/datasets/withmartian/routerbench/tree/784021482c3f320c6619ed4b3bb3b41a21424fcb}{RouterBench (revision \texttt{7840214})} and \href{https://huggingface.co/datasets/CARROT-LLM-Routing/SPROUT/tree/d54e3cda2189ca1ed13a877a44ed34736c4b6920}{SPROUT (revision \texttt{d54e3cd})}.}

RouterBench is restricted to the seven families listed above. Its released prompt parts are joined with two newline characters; SPROUT prompts are used as released. A query is eligible only if its prompt is nonempty and every candidate model has a finite reward in $[0,1]$ and a finite, nonnegative cost. We exclude queries exceeding 8,192 tokens under either main encoder's tokenizer, counting GTE's final EOS token. Exact duplicate prompt strings are removed across each benchmark, keeping the first eligible occurrence. Inputs are never truncated. We use both benchmarks' released rewards without rescaling. RouterBench costs use the released values. For SPROUT, the cost in USD per query is
\[
c(x,m)=\frac{n_{\mathrm{in}}(x,m)p_{\mathrm{in}}(m)+n_{\mathrm{out}}(x,m)p_{\mathrm{out}}(m)}{10^6},
\]
where $n_{\mathrm{in}}$ and $n_{\mathrm{out}}$ are the released input and output token counts, and $p_{\mathrm{in}}$ and $p_{\mathrm{out}}$ are the fixed prices in USD per million tokens listed in Table~\ref{tab:sprout-prices}. These prices come from the CARROT release.\footnote{\url{https://huggingface.co/CARROT-LLM-Routing}}

\begin{table}[H]
\setlength{\abovecaptionskip}{4pt}
\setlength{\belowcaptionskip}{6pt}
\centering
\small
\caption{Fixed SPROUT prices in USD per million tokens.}
\label{tab:sprout-prices}
\begin{tabular}{@{}lrr@{}}
\toprule
Model & Input & Output \\
\midrule
Claude-3.5-Sonnet-v1 & 3.00 & 15.00 \\
Titan Text Premier v1 & 0.50 & 1.50 \\
GPT-4o & 2.50 & 10.00 \\
GPT-4o-mini & 0.15 & 0.60 \\
Granite-3-2B Instruct & 0.10 & 0.10 \\
Granite-3-8B Instruct & 0.20 & 0.20 \\
Llama-3.1-8B Instruct & 0.20 & 0.20 \\
Llama-3.1-70B Instruct & 0.90 & 0.90 \\
Llama-3.2-1B Instruct & 0.06 & 0.06 \\
Llama-3.2-3B Instruct & 0.06 & 0.06 \\
Llama-3.3-70B Instruct & 0.90 & 0.90 \\
Llama-3-405B Instruct & 3.50 & 3.50 \\
Mixtral-8x7B-Instruct-v0.1 & 0.60 & 0.60 \\
\bottomrule
\end{tabular}
\end{table}

The retained datasets contain 34,771 RouterBench queries and 43,887 SPROUT queries. RouterBench uses 11 candidate models: WizardLM-13B-V1.2, Claude Instant v1, Claude v1, Claude v2, GPT-3.5-Turbo-1106, GPT-4-1106-preview, Code Llama Instruct 34B, Llama-2-70B-Chat, Mistral-7B-Chat, Mixtral-8x7B-Chat, and Yi-34B-Chat. SPROUT uses 13: Claude-3.5-Sonnet-v1, Titan Text Premier v1, GPT-4o, GPT-4o-mini, Granite-3-2B and 3-8B Instruct, Llama-3.1-8B and 70B Instruct, Llama-3.2-1B and 3B Instruct, Llama-3.3-70B Instruct, Llama-3-405B Instruct, and Mixtral-8x7B-Instruct-v0.1.

Family-cells use MMLU subjects and MT-Bench evaluation labels on RouterBench. On SPROUT, MATH subjects are recovered by matching question text after whitespace normalization; MMLU-Pro subjects are recovered using the released question indices with a text check. MuSR and RAGBench cells use released dataset identifiers. All other queries use the whole family as their cell, including OpenHermes and queries without a matched subject. Task and source identifiers are not interpreted as semantic subjects.

\paragraph{Splits and selection.}SPROUT's released training, validation and test partitions are pooled before preprocessing and constructing the evaluation splits below.
Five in-distribution splits reserve 30\% of each family for testing; 20\% of the outer training set is used for validation where required. Held-out evaluation has seven outer folds for RouterBench and six for SPROUT. Within each fold, validation holds out each remaining training family in turn, except for MLP and SharedTrunkNet, whose procedures are specified in Appendices~\ref{app:mlp-validation} and~\ref{app:sharedtrunknet}. Outside these exceptions, hyperparameters maximize mean routing utility over equally weighted validation queries. The test family is excluded from router-hyperparameter selection. The historical choice of the fixed L21 representation is described separately in Appendix~\ref{app:layer-sweep}.

\paragraph{Encoder inputs.}Both main encoders receive the prompts without an added instruction prefix or chat template. GTE uses its published bidirectional model and the final-layer state at the EOS token, appending EOS if it is not already present. Qwen2.5-7B-Instruct receives no added special tokens or EOS; L21 is the last nonpadding input token's state after transformer block 21, corresponding to \texttt{hidden\_states[21]}. Both encoders run in evaluation mode with bfloat16 weights. Extracted vectors are converted to float32 and unit-normalized before routing or geometric analysis; SharedTrunkNet uses the preprocessing specified in Appendix~\ref{app:sharedtrunknet}.\paragraph{Cost penalties.}
\[
\begin{aligned}
\text{RouterBench:}\quad &\{0,1,2,5,10,20,30,50,75,100,150,250,500\},\\
\text{SPROUT:}\quad &\{0,0.5,1,2,5,10,25,50,75,100,150,250,500\}.
\end{aligned}
\]

Mean utility across penalties is the arithmetic average of these thirteen evaluated values; it depends on this grid.

\paragraph{Simple routing rules.}
kNN estimates each model's reward and cost by uniformly averaging its recorded outcomes over the nearest training queries by cosine similarity, then selects the model with the highest estimated utility. kNN-100 fixes $k=100$. Validated kNN selects $k\in\{5,10,20,40,100,200\}$ per penalty. Family Adaptation uses multinomial logistic regression with balanced class weights, inverse regularization strength $C=1$ and at most 3,000 iterations, without coordinate standardization. Ridge predicts per-model utility with an unpenalized intercept and selects its regularization strength from $\{10^{-4},10^{-3},10^{-2},10^{-1},1,10,10^2,10^3,10^4\}$ by validation routing utility, separately for each penalty. It uses the same normalized vectors, without PCA. MLP training and validation are specified in Appendix~\ref{app:mlp-validation}.

We implement Avengers-Pro following \citet{zhang2025avengerspro} and KM-Prox following \citet{patel2025proxrouter}, using the utility objective and source-only validation protocol above. KM-Prox uses inverse temperature 20 in its fixed version and selects from $\{0,1,5,10,20,50,100\}$ in its validated version.

\paragraph{Aggregation and curves.}
In-distribution results average five split-level query means. Held-out results pool all test queries with equal query weights, so larger families contribute more to the mean. A displayed frontier retains only measured points that no other point of the same method dominates in both reward and cost. No interpolation is used for utility comparisons. The tables retain all penalties, including dominated points. Cost axes use a symmetric logarithmic scale. Routing confidence intervals are defined below.

\paragraph{Confidence intervals.}\label{app:routing-ci}
Where reported, routing confidence intervals are pointwise 95\% percentile bootstrap intervals~\citep{efron1979bootstrap} from 2,000 paired samples of test queries. In OOD evaluation, queries are resampled within each held-out family, preserving family sizes. In in-distribution evaluation, we resample unique test queries and keep their repeated appearances across splits together. Each appearance retains its weight in the original five-split mean, with weights renormalized in each sample. The same samples are used across methods and penalties; mean-utility intervals are computed after averaging the thirteen penalties within each sample. Intervals are conditional on the evaluated families, splits and fitted routers, and do not include retraining variability or uncertainty over new families. The lookup-disagreement analysis uses 1,000 paired samples, as specified in Appendix~\ref{app:iid-lookup-disagreement}.

\subsection{MLP training and validation}\label{app:mlp-validation}
For each cost penalty, MLP fits a separate scalar regressor for each candidate model to its recorded utility $r(x,m)-\lambda c(x,m)$. It selects the model with the largest predicted utility. Inputs are the unit-normalized GTE or L21 embeddings, without PCA or coordinate standardization; output values are unrestricted. Table~\ref{tab:mlp-settings} gives the training settings.

\begin{table}[H]
\setlength{\abovecaptionskip}{4pt}
\setlength{\belowcaptionskip}{6pt}
\centering\small
\caption{MLP candidate grid and fixed training settings.}
\label{tab:mlp-settings}
\begin{tabular}{@{}ll@{}}
\toprule
Setting & Values \\
\midrule
Hidden layers & $(100,100)$ or $(100,100,100)$, with ReLU \\
L2 regularization $\alpha$ & $10^{-4},\;10^{-2},\;1,\;10,\;100$ \\
Training objective & Squared error with L2 weight regularization \\
Optimizer & Adam; learning rate $10^{-3}$ \\
Adam parameters & $\beta_1=0.9$, $\beta_2=0.999$, $\epsilon=10^{-8}$ \\
Batch size / maximum epochs & 200 / 200 \\
Early stopping & Enabled; internal validation fraction $0.1$ \\
Stopping criterion & Validation $R^2$; tolerance $10^{-4}$, patience parameter 10 \\
\bottomrule
\end{tabular}
\end{table}

The two architectures and five regularization values give ten candidate configurations at each of thirteen penalties: 130 candidate settings per outer evaluation. A configuration is shared across the candidate-model regressors, but their weights are trained separately. We select the configuration by the realized utility of its routing decisions on validation queries, separately for each benchmark, encoder, outer split and penalty.

For ID evaluation, selection uses the 20\% validation split of the outer training data. For held-out-family evaluation, it uses two predetermined source families: each is held out in turn while fitting on the other source families, and validation utilities are pooled with equal query weights. The two validation families are fixed before fitting and are identical for GTE and L21. The outer test family is excluded throughout.

After selection, only the winning configuration is refitted on the outer training set. Every fit, including the final refit, reserves its own random 10\% of fitting queries for early stopping and restores the weights with the highest validation $R^2$; this internal subset is separate from the queries used to select the routing configuration. The same procedure is used for both encoders.

\subsection{SharedTrunkNet training and validation}
\label{app:sharedtrunknet}

\paragraph{Comparison and inputs.}
We adapt the released SharedTrunkNet pipeline of \citet{varshney2026prefill} to predict utility $u_\lambda(x,m)=r(x,m)-\lambda c(x,m)$.
All compared methods use the same raw prompts, frozen Qwen2.5-7B-Instruct checkpoint, candidate models, recorded outcomes, outer splits and penalty grids.
kNN-100 and Family Adaptation use unit-normalized L21 states as in the main text.
SharedTrunkNet retains its own feature preprocessing and network.
The comparison evaluates these routing pipelines with a common encoder and utility objective; it does not reproduce the original paper's model pools or absolute scores.

\paragraph{Features and layer selection.}
For each layer $\ell\in\{15,\ldots,28\}$, we use the last input token's hidden state.
We standardize its coordinates and fit an unwhitened PCA with 100 components on the outer training queries only.
These transforms remain fixed during feature selection.
For each candidate model and penalty, a Ridge probe with fixed $\alpha=1$ and an intercept predicts utility in five shuffled folds of the outer training queries.
We apply the released ternary layer-search rule to negative mean validation MSE.
This selects a layer separately for every model and penalty; it is a heuristic search, not an exhaustive maximization.
The selected 100-dimensional blocks are concatenated in candidate-model order, giving $100M$ inputs for $M$ candidate models.
Blocks are repeated when models select the same layer.
The five folds split training queries, not whole families; the outer test family is excluded from all preprocessing, selection and fitting.

\paragraph{Network and utility objective.}
We use the released shared network: hidden widths 256 and 128, ReLU activations, dropout 0.3 and 0.2, and $M$ linear outputs.
Training uses Adam with learning rate $10^{-3}$, weight decay $10^{-4}$, batches of 512, at most 150 epochs and early stopping with patience 15 and improvement threshold $10^{-4}$.
Each of ten independently initialized networks reserves a random 15\% of the outer training queries for early stopping; the five networks with lowest validation MSE are retained.
Their raw outputs are averaged, and the model with the largest predicted utility is selected.
We train a separate ensemble for each penalty.
The utility adaptation replaces logistic-regression/AUC feature selection with Ridge/MSE, and binary cross-entropy with utility MSE in the network.
Outputs are not passed through a sigmoid, and cost is not subtracted again.
The architecture, optimizer, stopping rule and ten-to-five ensemble procedure are retained.

\paragraph{Alternative selection procedures.}
The fixed-L21 control repeats the L21 PCA block for every candidate model.
The common-layer control selects one layer for all models and penalties by mean validation routing utility over the thirteen penalties.
It trains one screening network per layer and penalty using the two source-family holdouts defined for MLP in Appendix~\ref{app:mlp-validation}, fitting preprocessing within each training fold.
Both controls use the same final shared architecture and ten-to-five ensemble rule, with the common layer's block repeated $M$ times.
Thus, these controls compare selection procedures and feature assembly.

\paragraph{Other reproducibility information.}\label{app:other-reproducibility}
ID outer splits use seeds 0--4 and their inner validation splits use 10000--10004, respectively. MLP uses seed 1234 for initialization, internal early-stopping splits and batch shuffling; its two source-family validation folds are fixed before fitting, identically for both encoders.\footnote{For each held-out family, the two MLP validation families are: RouterBench: ARC or GSM8K $\to$ HellaSwag, MMLU; HellaSwag $\to$ MT-Bench, GSM8K; MBPP $\to$ WinoGrande, HellaSwag; MMLU $\to$ WinoGrande, MBPP; MT-Bench $\to$ HellaSwag, MBPP; WinoGrande $\to$ ARC, MMLU. SPROUT: GPQA $\to$ MMLU-Pro, OpenHermes; MATH $\to$ MMLU-Pro, MuSR; MMLU-Pro $\to$ GPQA, RAGBench; MuSR $\to$ OpenHermes, RAGBench; OpenHermes $\to$ MuSR, MMLU-Pro; RAGBench $\to$ OpenHermes, MATH.} Avengers-Pro jointly selects the number of clusters from $\{32,64,128\}$, retrieved clusters from $\{1,3\}$ and reward--cost mixing weight from $\{0,0.01,\ldots,1\}$ at each penalty, with cluster-weighting parameter $\beta=9$. KM-Prox uses 32 clusters. Both use ordinary k-means with ten initializations and seed 42. SharedTrunkNet uses seed 42 for PCA, probe-validation folds and common-layer screening, and seeds 42--51 for ensemble members. Geometric sampling, PCA, t-SNE and k-means use seed 42; t-SNE uses PCA initialization, automatic learning rate and 1,000 iterations. Routing and cost--reward bootstrap analyses use seed 20261005; lookup-disagreement intervals use seed 42. The software environment uses Python 3.11.16, PyTorch 2.5.1, Transformers 4.44.2, NumPy 1.26.4 and scikit-learn 1.5.2. Encoder revisions are GTE \texttt{a8d08b3}, Qwen \texttt{a09a354} and MiniLM \texttt{a50ef00}.

\subsection{Computing infrastructure and asset documentation}
\label{app:resources-assets}

\paragraph{Computing infrastructure.}
Experiments ran on a shared Linux cluster managed by Slurm, with AMD EPYC and Intel Xeon CPUs. Recorded GPU hardware includes NVIDIA A100 80GB and H100 80GB devices. Neural-router workers requested one GPU and four allocated CPUs, with Slurm host-memory requests of \texttt{32G} for MLP and \texttt{24G} for SharedTrunkNet. CPU baseline jobs requested eight allocated CPUs and \texttt{48G} host memory; PCA preprocessing used sixteen allocated CPUs. These are job allocations, not minimum hardware requirements. Software versions are specified above.

\paragraph{Licenses and access.}
The \href{https://huggingface.co/Alibaba-NLP/gte-Qwen2-7B-instruct}{GTE}, \href{https://huggingface.co/Qwen/Qwen2.5-7B-Instruct}{Qwen} and \href{https://huggingface.co/sentence-transformers/all-MiniLM-L12-v2}{MiniLM} checkpoints and the reused \href{https://github.com/NVIDIA-AI-Blueprints/llm-router/blob/8a9d3509fbde879d9795258081bab9553b458e04/LICENSE}{NVIDIA router code} use Apache 2.0; \href{https://github.com/withmartian/routerbench/blob/main/LICENSE}{RouterBench software} uses MIT. The \href{https://huggingface.co/datasets/withmartian/routerbench}{RouterBench} and \href{https://huggingface.co/datasets/CARROT-LLM-Routing/SPROUT}{SPROUT} dataset cards do not specify a separate dataset license. Software licenses do not establish the terms of all underlying prompts or generated responses.

The constituent dataset cards declare MIT for \href{https://huggingface.co/datasets/cais/mmlu}{MMLU}, \href{https://huggingface.co/datasets/Rowan/hellaswag}{HellaSwag}, \href{https://huggingface.co/datasets/openai/gsm8k}{GSM8K} and \href{https://huggingface.co/datasets/TIGER-Lab/MMLU-Pro}{MMLU-Pro}; CC BY-SA 4.0 for \href{https://huggingface.co/datasets/allenai/ai2_arc}{ARC}; and CC BY 4.0 for \href{https://huggingface.co/datasets/google-research-datasets/mbpp}{MBPP}, \href{https://huggingface.co/datasets/Idavidrein/gpqa}{GPQA}, \href{https://huggingface.co/datasets/TAUR-Lab/MuSR}{MuSR} and \href{https://huggingface.co/datasets/galileo-ai/ragbench}{RAGBench}. The \href{https://github.com/hendrycks/math/blob/main/LICENSE}{MATH} and \href{https://github.com/allenai/winogrande/blob/master/LICENSE}{WinoGrande} repositories publish MIT and Apache 2.0 software licenses, respectively. We do not infer additional question-level permissions from these, from the license of a separate MT-Bench annotation release, or from the availability of \href{https://huggingface.co/datasets/teknium/OpenHermes-2.5}{OpenHermes}, whose card does not specify a license. The \href{https://github.com/ZhangYiqun018/AvengersPro}{Avengers-Pro README} declares MIT, although its linked license file was unavailable; we did not identify a repository-wide license in the consulted \href{https://github.com/ynulihao/LLMRouterBench}{LLMRouterBench} release. Applicable upstream terms remain in force. The anonymous package will preserve required third-party notices and provide download and preparation instructions rather than redistribute benchmark text or pretrained weights. This also respects GPQA's request not to publish question examples online.

\paragraph{Consent information.}
We reuse released benchmarks and collect no new data from human participants. The releases do not provide consent information for every record. OpenHermes lists LMSYS-Chat-1M among its sources; the \href{https://huggingface.co/datasets/lmsys/lmsys-chat-1m}{LMSYS documentation} reports consent through the collection website's terms of use. This does not establish the consent status of every OpenHermes example in SPROUT, whose individual upstream sources were not audited here.

\paragraph{Potentially sensitive content.}
The reused text may contain personally identifying or offensive content. For example, LMSYS reports efforts to remove identifying information but also retains unsafe conversations. We do not claim that every benchmark example is free of such content or that we performed a comprehensive content audit. The package will link to the original data releases instead of republishing their raw prompts and responses.
  \section{Geometric Measurements}\label{app:geometry}
\subsection{Neighborhood measures and visualization}
Neighbor comparisons use original unit-normalized embeddings. Same-family share is the fraction of retrieved examples from the query's family; its random comparison is the fraction of eligible reference queries from that family, averaged over queries. Largest-family share is the largest fraction of a query's neighbors belonging to any one family, averaged over queries.

The three-panel visualization uses a fixed t-SNE projection~\citep{vandermaaten2008tsne} of each displayed set, after PCA to at most 50 dimensions. It samples at most 600 queries per family, at most 400 per displayed MMLU subject, and at most 1,200 in professional law. The middle panel uses the five largest MMLU subjects by query count; the last panel uses professional law, the largest subject. Each panel uses perplexity 30 and a fixed random state. These are descriptive projections of the dataset, not estimates of test performance.

\subsection{Difficulty within family-cells}\label{app:difficulty-locality}
For the difficulty comparison, we use the solve fraction: the fraction of candidate models achieving reward at least 0.5 on a query. Higher solve fractions indicate easier queries. We use GTE embeddings and the five ID splits. For each test query, we retrieve its ten nearest training queries from the same family-cell and average their absolute differences in solve fraction from the query. The random comparison averages this difference over all training queries in the same cell. Cells with fewer than ten eligible training queries are excluded. Averaging both quantities equally over eligible test-query appearances across the five splits gives $D_{\mathrm{neighbors}}$ and $D_{\mathrm{random}}$. Table~\ref{tab:difficulty-locality} reports these differences and their relative reduction, $100(1-D_{\mathrm{neighbors}}/D_{\mathrm{random}})\%$.

\begin{table}[H]
\setlength{\abovecaptionskip}{4pt}
\setlength{\belowcaptionskip}{6pt}
\centering\small
\caption{Mean absolute difference in solve fraction within family-cells.}
\label{tab:difficulty-locality}
\begin{tabular}{@{}lrrr@{}}
\toprule
Benchmark & Neighbors & Random selection & Relative reduction \\
\midrule
RouterBench & 0.2463 & 0.2565 & 4.01\% \\
SPROUT & 0.1861 & 0.2467 & 24.58\% \\
\bottomrule
\end{tabular}
\end{table}

These results measure similarity in observed difficulty, not the utility of a routing decision.

On SPROUT, the additional similarity may partly reflect finer groups not identified by the available labels. If these groups differ in average solve fraction and are separated in the embedding space, neighbors can have similar difficulty without the space being organized directly by difficulty.

\subsection{Family and subject structure}\label{app:family-topic-geometry}

For each test query, we measure the share of its 100 nearest training queries belonging to its family. Table~\ref{tab:full-gte-family-locality} reports full-data GTE results, averaged over queries and the five ID splits. Random selection samples uniformly from all training queries.

\begin{table}[H]
\setlength{\abovecaptionskip}{4pt}
\setlength{\belowcaptionskip}{6pt}
\centering\small
\caption{Full-data GTE same-family share among 100 neighbors.}
\label{tab:full-gte-family-locality}
\begin{tabular}{@{}lrr@{}}
\toprule
Benchmark & Nearest neighbors & Random selection \\
\midrule
RouterBench & 95.51\% & 29.55\% \\
SPROUT & 97.30\% & 31.88\% \\
\bottomrule
\end{tabular}
\end{table}

We test subject organization in the original normalized embedding space, separately within each family with at least two known subjects. On each of the five ID splits, ordinary k-means is fitted to training embeddings with one cluster per observed training subject and ten initializations. Test queries are assigned to the nearest fitted center. Subject labels determine the number of clusters and are used to score the assignments; they do not enter the clustering objective. We report normalized mutual information (NMI; larger values indicate greater agreement)~\citep{vinh2010information} and adjusted Rand index (ARI; adjusted for chance agreement)~\citep{hubert1985comparing}, averaged over the five splits. Missing subject labels are excluded. Table~\ref{tab:subject-recovery} covers all eligible families with native subject labels: MMLU on RouterBench, and MATH and MMLU-Pro on SPROUT. Native task or source identifiers are not treated as subjects.

We repeat the locality and subject-recovery tests with \texttt{sentence-transformers/all-MiniLM-L12-v2}~\citep{sentencetransformers_minilm_l12_v2,reimers2019sentencebert,wang2020minilm}, using attention-mask mean pooling. Prompts exceeding its 512-token limit, including special tokens, are excluded rather than truncated. Both encoders use the same retained queries and splits: 34,659 RouterBench queries and 41,324 SPROUT queries. All MuSR prompts exceed this limit, leaving five SPROUT families.

\begin{table}[H]
\setlength{\abovecaptionskip}{4pt}
\setlength{\belowcaptionskip}{6pt}
\centering\small
\setlength{\tabcolsep}{5pt}
\caption{Recovery of native subjects within families on test queries. Matched columns use the same MiniLM-compatible queries and splits.}
\label{tab:subject-recovery}
\begin{tabular}{@{}llrrrrr@{}}
\toprule
& & & \multicolumn{2}{c}{Full-data GTE} & \multicolumn{2}{c}{Matched subset NMI} \\
\cmidrule(lr){4-5}\cmidrule(lr){6-7}
Benchmark & Family & Subjects & NMI & ARI & GTE & MiniLM \\
\midrule
RouterBench & MMLU & 57 & 0.722 & 0.418 & 0.726 & 0.640 \\
SPROUT & MATH & 7 & 0.327 & 0.246 & 0.334 & 0.296 \\
SPROUT & MMLU-Pro & 14 & 0.286 & 0.172 & 0.280 & 0.505 \\
\bottomrule
\end{tabular}
\end{table}

For each test query, we measure the same-family share among its 100 nearest training examples by cosine similarity, then average over queries and the five splits. Both encoders retain strong family locality (Table~\ref{tab:semantic-family-locality}); subject recovery, measured by the k-means procedure above, varies by family (Table~\ref{tab:subject-recovery}).

\begin{table}[H]
\setlength{\abovecaptionskip}{4pt}
\setlength{\belowcaptionskip}{6pt}
\centering\small
\setlength{\tabcolsep}{6pt}
\caption{Same-family share among 100 nearest training queries, on the matched subset. Random selection samples uniformly from all training queries.}
\label{tab:semantic-family-locality}
\begin{tabular}{@{}lrrr@{}}
\toprule
Benchmark & GTE & MiniLM & Random selection \\
\midrule
RouterBench & 95.48\% & 90.43\% & 29.49\% \\
SPROUT & 97.33\% & 90.74\% & 34.53\% \\
\bottomrule
\end{tabular}
\end{table}

\subsection{Family concentration in held-out neighborhoods}
\label{app:family-concentration}
For each query, we exclude its entire family and retrieve the 100 nearest remaining queries by cosine similarity. We measure the largest fraction of these neighbors belonging to any one training family, then average this fraction equally over all queries. GTE and L21 use the same queries and eligible training pools.

\begin{table}[H]
\setlength{\abovecaptionskip}{4pt}
\setlength{\belowcaptionskip}{6pt}
\centering\small
\caption{Mean largest-family share among 100 neighbors, with the query's family excluded.}
\label{tab:family-concentration}
\begin{tabular}{@{}lrr@{}}
\toprule
Benchmark & GTE & L21 \\
\midrule
RouterBench & 78.4\% & 72.7\% \\
SPROUT & 84.1\% & 92.0\% \\
\bottomrule
\end{tabular}
\end{table}

\subsection{Answer-format agreement in held-out neighborhoods}\label{app:answer-format}
We test whether changing the encoder changes the answer formats represented among a query's nearest neighbors. We use two groups, assigned from the benchmark's task families. On RouterBench, ARC, HellaSwag, MMLU and WinoGrande are multiple-choice; GSM8K, MBPP and MT-Bench are free-form. On SPROUT, GPQA, MMLU-Pro and MuSR are multiple-choice; MATH, OpenHermes and RAGBench are free-form. These are coarse task labels, not classifications of the models' generated answers.

For each query, we exclude its entire family and retrieve the 100 nearest remaining queries by cosine similarity. We measure the fraction whose answer-format label matches the query's. We use identical queries and eligible neighbors for GTE and L21, and average over all held-out queries with equal weights. The random-selection column is the expected matching fraction when sampling uniformly from the same eligible queries. The analysis covers all seven RouterBench families (34,771 queries) and all six SPROUT families (43,887 queries).

\begin{table}[H]
\setlength{\abovecaptionskip}{4pt}
\setlength{\belowcaptionskip}{6pt}
\centering
\small
\caption{Percentage of held-out neighbors sharing the query's answer-format label.}
\label{tab:answer-format}
\begin{tabular}{@{}lrrr@{}}
\toprule
Benchmark & Randomly selected queries & GTE & L21 \\
\midrule
RouterBench & 51.2\% & 76.4\% & 99.9\% \\
SPROUT & 38.9\% & 58.9\% & 90.2\% \\
\bottomrule
\end{tabular}
\end{table}
L21 neighborhoods follow this answer-format grouping much more closely, even when the query's own family is unavailable. This measurement establishes a change in neighborhood composition; its contribution to routing utility is a separate question.

\subsection{Geometric Proximity and Model Performance}\label{app:transfer}
We test whether the family supplying the most neighbors also best predicts utility differences between candidate models.
For each held-out family and each remaining training family, we retrieve the nearest 100 queries within that training family using GTE cosine similarity. If the family has fewer than 100 queries, we use all of them; this applies to MT-Bench, which has 80 queries. We average each candidate model's recorded rewards and costs over these neighbors, giving predicted utility $\widehat u_\lambda(x,m)=\widehat r(x,m)-\lambda\widehat c(x,m)$ at every prescribed cost penalty.

We center both predicted and observed utilities across the $M$ candidate models, separately for each query:
\[
\widetilde u_\lambda(x,m)
= u_\lambda(x,m)-\frac{1}{M}\sum_{j=1}^{M}u_\lambda(x,m_j).
\]
For each training family, we measure the mean squared error between its predicted and observed centered utilities, averaged over models and queries in the held-out family. This tests relative model preferences. The family with the lowest error is identified retrospectively for this diagnostic.

We then rank the training families by their shares of the unrestricted 100-neighbor set, averaged over queries in the held-out family. Rank 1 means the largest share. Table~\ref{tab:utility-family-rank} reports the retrieval rank of the lowest-error family and how often that family ranks first. Exact ties receive average ranks; results are averaged over tied lowest-error families. Mean ranks weight held-out families equally at each penalty. The final row averages these ranks equally across the thirteen penalties.

\begin{table}[H]
\setlength{\abovecaptionskip}{4pt}
\setlength{\belowcaptionskip}{6pt}
\centering\small
\setlength{\tabcolsep}{10pt}
\caption{Retrieval rank of the family with the lowest relative-utility prediction error. Top-1 counts held-out families for which it is also the most retrieved. Each benchmark uses its prescribed penalty grid.}
\label{tab:utility-family-rank}
\begin{tabular}{@{}rrrrr@{}}
\toprule
& \multicolumn{2}{c}{RouterBench (six sources)} & \multicolumn{2}{c}{SPROUT (five sources)} \\
\cmidrule(lr){2-3}\cmidrule(lr){4-5}
$\lambda$ & Mean rank & Top-1 & Mean rank & Top-1 \\
\midrule
0 & 3.14 & 0/7 & 3.00 & 2/6 \\
0.5 & n/a & n/a & 3.00 & 2/6 \\
1 & 3.14 & 0/7 & 3.00 & 2/6 \\
2 & 3.14 & 0/7 & 3.00 & 2/6 \\
5 & 3.36 & 0/7 & 3.00 & 2/6 \\
10 & 3.36 & 0/7 & 3.00 & 2/6 \\
20 & 2.71 & 1/7 & n/a & n/a \\
25 & n/a & n/a & 3.33 & 1/6 \\
30 & 3.29 & 1/7 & n/a & n/a \\
50 & 3.29 & 1/7 & 3.33 & 1/6 \\
75 & 3.29 & 1/7 & 3.33 & 1/6 \\
100 & 3.29 & 1/7 & 3.67 & 1/6 \\
150 & 3.29 & 1/7 & 3.67 & 1/6 \\
250 & 3.29 & 1/7 & 3.17 & 2/6 \\
500 & 3.57 & 0/7 & 3.17 & 2/6 \\
\midrule
Mean / total & 3.24 & 7/91 & 3.21 & 21/78 \\
\bottomrule
\end{tabular}
\end{table}

\subsection{Predicting Model Preferences}\label{app:ordering}
For every held-out query, we average recorded rewards and costs over its 100 nearest training examples. At each penalty, we compare the sign of each predicted pairwise utility difference with the recorded difference. Recorded ties are excluded; predicted ties receive half credit. Accuracy is first averaged within each query, then over queries with at least one non-tied pair. This separates the accuracy of model comparisons from the utility of the final selected model.

\begin{figure}[H]
\centering
\includegraphics[width=.9\linewidth]{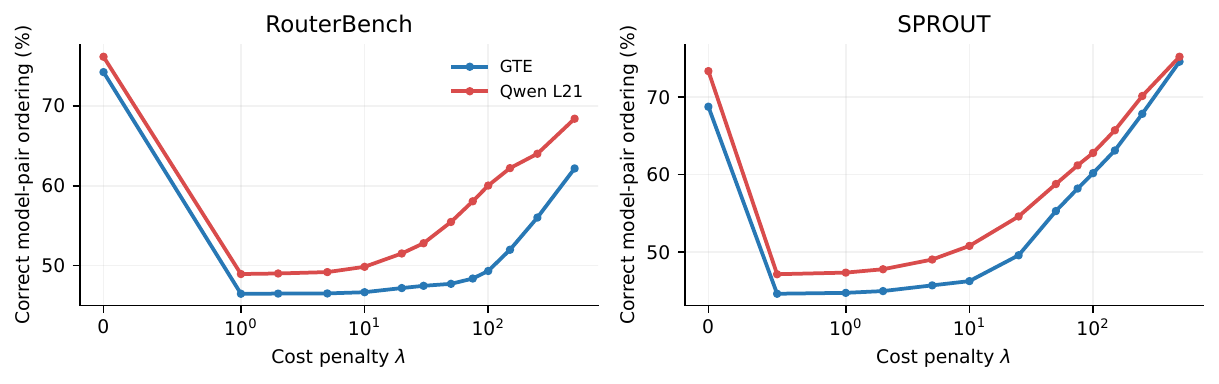}
\caption{Correct utility ordering across candidate-model pairs, evaluated at every penalty. At $\lambda=0$, recorded reward ties change the set of evaluated pairs.}
\end{figure}

\begin{figure}[H]
\centering
\includegraphics[width=.9\linewidth]{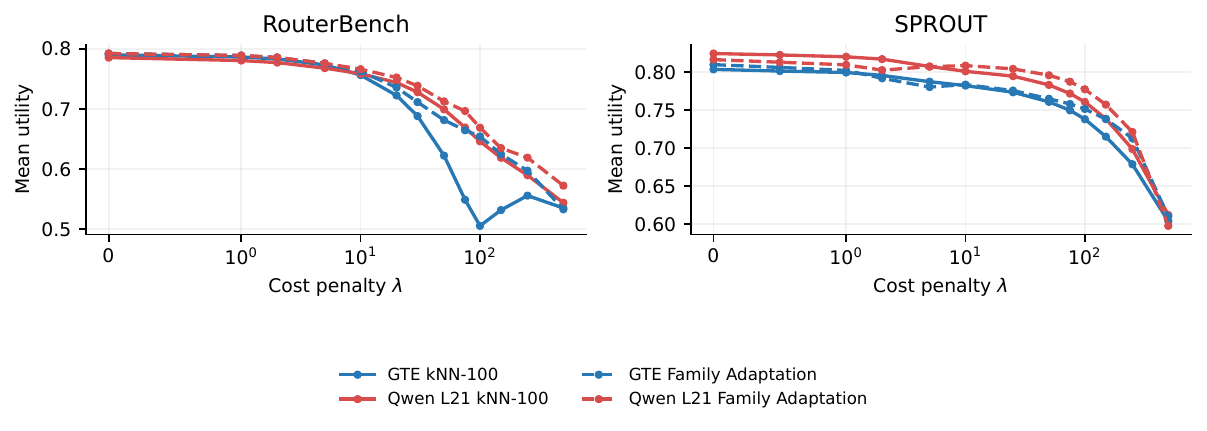}
\caption{Held-out utility at every penalty for kNN-100 and Family Adaptation. Unlike reward--cost frontiers, these curves retain every evaluated policy.}
\end{figure}

\clearpage
\section{ID Routing and Group Lookups}
\label{app:iid-routing-lookups}

\subsection{Query-Specific Decisions Beyond Group Lookup}
\label{app:iid-lookup-disagreement}
\paragraph{Question and comparison.}
Similar overall scores do not tell us whether kNN makes useful choices beyond group lookup. We compare GTE kNN-100 with family lookup and family-cell lookup on the same five ID splits of each benchmark, at all thirteen cost penalties. Family lookup selects the model with the highest mean training utility within the query's family. Family-cell lookup uses the available native labels within that family; where no finer label exists, the cell is the family. A cell absent from training falls back to family lookup. kNN-100 instead selects the model with the highest mean utility among the query's 100 nearest training examples. All choices use training outcomes only.

\paragraph{What we measure.}
For each test query, we record whether kNN and the lookup choose different models. We then subtract the recorded utility of the lookup's choice from that of kNN's choice. We report the fraction of different choices, the mean utility gain on those choices, and the gain over all test queries. Agreeing choices contribute zero, so the overall gain equals the disagreement fraction multiplied by the gain on disagreements. We average queries equally within each split, then average the five splits equally. Table~\ref{tab:iid-disagreement-summary} also averages the thirteen penalties equally; its gain on disagreements is the overall mean gain divided by the mean disagreement fraction.

\paragraph{Uncertainty.}
The 95\% intervals use 1,000 paired bootstrap samples of unique test queries, keeping each query's repeated test appearances and all its penalty-specific results together. These intervals describe uncertainty over test queries for the fixed splits and fitted routers; they do not include retraining variability. Per-penalty intervals are pointwise.

\CIDisagreementSummary

\paragraph{Results.}
The additional gain is small on both benchmarks. Relative to family-cell lookup, kNN improves mean utility by 0.0054 on RouterBench and 0.0065 on SPROUT, despite choosing a different model in 28.78\% and 47.75\% of cases, respectively. Thus, its query-specific decisions contain some useful information beyond a fixed choice per group, but this adds only a small average improvement. The effect varies with the cost penalty (Table~\ref{tab:iid-disagreement-penalties}): on RouterBench the intervals include zero at $\lambda=0,1,2$, while SPROUT has positive pointwise intervals throughout and a larger gain of 0.0228 at $\lambda=500$.

\CIDisagreementPenalties

\begin{samepage}
\subsection{Lookup with predicted group labels}\label{app:lookup-deployability}
This check replaces the known family or cell label with a label predicted from the query embedding. On each of the five ID splits, we fit separate multinomial logistic classifiers for family and native-cell labels using GTE embeddings. Coordinate standardization is fitted on training queries only. Both classifiers use fixed $C=1$, unweighted classes and the L-BFGS solver, with tolerance $10^{-4}$ and at most 3,000 iterations; there is no hyperparameter search. We then select the model with the highest mean training utility in the predicted group. All nonempty training cells are retained, matching the known-label lookup. No test group label is supplied to the predicted-label router.\par
\end{samepage}

We evaluate both benchmarks at all thirteen penalties, reporting label accuracy and routing utility alongside the corresponding known-label lookup. Each result is averaged over the five splits.

Table~\ref{tab:lookup-deployability} summarizes all thirteen penalties. Predicting labels preserves almost all of the known-label lookup's mean utility on both benchmarks. Cell labels are less accurately predicted than family labels, but their mean utility loss remains below $0.0006$.

\CIDeployability

\clearpage
\section{Routing Results Across All Penalties}\label{app:results}
\subsection{Overall routing results}
ID results average five splits; OOD results pool queries from all held-out families. Native-label lookups are reported for ID evaluation. The Mean column averages all thirteen penalties and is repeated in each part of the full tables. The realized oracle selects, for each test query and cost penalty, the model with the highest recorded utility.

In what follows, the tables report:
\begin{itemize}[leftmargin=1.2em,topsep=0.25em,itemsep=0.15em,parsep=0pt]
\item \textbf{Table~\ref{tab:ood-mean}:} mean OOD utility across all thirteen cost penalties for the GTE methods, the best single model, and our two L21 routers.
\item \textbf{Tables~\ref{tab:ci-routerbench-iid-1}--\ref{tab:ci-routerbench-iid-3}:} RouterBench ID results at every cost penalty.
\item \textbf{Tables~\ref{tab:ci-routerbench-ood-1}--\ref{tab:ci-routerbench-ood-3}:} RouterBench OOD results at every cost penalty.
\item \textbf{Tables~\ref{tab:ci-sprout-iid-1}--\ref{tab:ci-sprout-iid-3}:} SPROUT ID results at every cost penalty.
\item \textbf{Tables~\ref{tab:ci-sprout-ood-1}--\ref{tab:ci-sprout-ood-3}:} SPROUT OOD results at every cost penalty.
\item \textbf{Table~\ref{tab:nvidia-mean-ci}:} mean OOD utility for SharedTrunkNet with fixed L21, a common selected layer, or per-model selected layers, alongside our L21 routers.
\item \textbf{Table~\ref{tab:sharedtrunknet-differences}:} paired utility differences between each L21 router and SharedTrunkNet with per-model selected layers.
\item \textbf{Tables~\ref{tab:sharedtrunknet-routerbench}--\ref{tab:sharedtrunknet-sprout}:} the same comparison at every cost penalty on RouterBench and SPROUT, respectively.
\item \textbf{Tables~\ref{tab:family-routing-routerbench}--\ref{tab:family-routing-sprout}:} the change from GTE to L21 for kNN-100, broken down by held-out family on RouterBench and SPROUT, with both query-weighted and family-weighted averages.
\end{itemize}
All tables include 95\% confidence intervals.

\CIOodMean

\begin{table}[H]
\setlength{\abovecaptionskip}{4pt}
\setlength{\belowcaptionskip}{6pt}
\centering\small
\setlength{\tabcolsep}{3pt}
\caption{RouterBench in-distribution utility with 95\% intervals, part 1 of 3. Mean averages all thirteen penalties.}
\label{tab:ci-routerbench-iid-1}
\begin{tabular}{@{}lrrrrrr@{}}
\toprule
Method / $\lambda$ & 0 & 1 & 2 & 5 & 10 & Mean \\[2pt]
\midrule
\resultlabel{GTE / Avengers-Pro} & \resultci{0.7913}{0.7869}{0.7957} & \resultci{0.7888}{0.7844}{0.7932} & \resultci{0.7863}{0.7819}{0.7907} & \resultci{0.7788}{0.7744}{0.7833} & \resultci{0.7664}{0.7618}{0.7709} & \resultci{0.7176}{0.7134}{0.7219} \\[2pt]
\resultlabel{GTE / Best single model} & \resultci{0.7922}{0.7879}{0.7967} & \resultci{0.7890}{0.7846}{0.7934} & \resultci{0.7857}{0.7813}{0.7902} & \resultci{0.7759}{0.7715}{0.7805} & \resultci{0.7596}{0.7550}{0.7641} & \resultci{0.6994}{0.6951}{0.7035} \\[2pt]
\resultlabel{GTE / Family lookup} & \resultci{0.7925}{0.7880}{0.7969} & \resultci{0.7895}{0.7850}{0.7939} & \resultci{0.7867}{0.7822}{0.7912} & \resultci{0.7785}{0.7739}{0.7830} & \resultci{0.7708}{0.7661}{0.7753} & \resultci{0.7212}{0.7172}{0.7252} \\[2pt]
\resultlabel{GTE / Family-cell lookup} & \resultci{0.7916}{0.7870}{0.7960} & \resultci{0.7884}{0.7839}{0.7929} & \resultci{0.7855}{0.7810}{0.7900} & \resultci{0.7775}{0.7730}{0.7820} & \resultci{0.7697}{0.7650}{0.7742} & \resultci{0.7229}{0.7187}{0.7271} \\[2pt]
\resultlabel{GTE / KM-Prox} & \resultci{0.7936}{0.7892}{0.7980} & \resultci{0.7906}{0.7861}{0.7950} & \resultci{0.7876}{0.7831}{0.7920} & \resultci{0.7799}{0.7754}{0.7843} & \resultci{0.7733}{0.7687}{0.7779} & \resultci{0.7287}{0.7246}{0.7329} \\[2pt]
\resultlabel{GTE / KM-Prox (validated)} & \resultci{0.7949}{0.7904}{0.7993} & \resultci{0.7914}{0.7870}{0.7960} & \resultci{0.7884}{0.7840}{0.7928} & \resultci{0.7807}{0.7762}{0.7852} & \resultci{0.7735}{0.7690}{0.7780} & \resultci{0.7289}{0.7248}{0.7330} \\[2pt]
\resultlabel{GTE / MLP} & \resultci{0.7934}{0.7889}{0.7978} & \resultci{0.7876}{0.7832}{0.7919} & \resultci{0.7859}{0.7816}{0.7904} & \resultci{0.7764}{0.7721}{0.7809} & \resultci{0.7668}{0.7623}{0.7710} & \resultci{0.7245}{0.7205}{0.7285} \\
\bottomrule
\end{tabular}
\end{table}

\begin{table}[H]
\setlength{\abovecaptionskip}{4pt}
\setlength{\belowcaptionskip}{6pt}
\centering\small
\setlength{\tabcolsep}{3pt}
\caption{RouterBench in-distribution utility with 95\% intervals, part 2 of 3. Mean averages all thirteen penalties.}
\label{tab:ci-routerbench-iid-2}
\begin{tabular}{@{}lrrrrr@{}}
\toprule
Method / $\lambda$ & 20 & 30 & 50 & 75 & Mean \\[2pt]
\midrule
\resultlabel{GTE / Avengers-Pro} & \resultci{0.7407}{0.7360}{0.7453} & \resultci{0.7188}{0.7139}{0.7237} & \resultci{0.7022}{0.6974}{0.7071} & \resultci{0.6937}{0.6887}{0.6987} & \resultci{0.7176}{0.7134}{0.7219} \\[2pt]
\resultlabel{GTE / Best single model} & \resultci{0.7269}{0.7221}{0.7316} & \resultci{0.6942}{0.6893}{0.6991} & \resultci{0.6537}{0.6482}{0.6591} & \resultci{0.6490}{0.6435}{0.6545} & \resultci{0.6994}{0.6951}{0.7035} \\[2pt]
\resultlabel{GTE / Family lookup} & \resultci{0.7568}{0.7522}{0.7614} & \resultci{0.7429}{0.7384}{0.7474} & \resultci{0.7157}{0.7109}{0.7204} & \resultci{0.6958}{0.6908}{0.7006} & \resultci{0.7212}{0.7172}{0.7252} \\[2pt]
\resultlabel{GTE / Family-cell lookup} & \resultci{0.7561}{0.7515}{0.7608} & \resultci{0.7424}{0.7378}{0.7470} & \resultci{0.7172}{0.7123}{0.7219} & \resultci{0.6993}{0.6941}{0.7043} & \resultci{0.7229}{0.7187}{0.7271} \\[2pt]
\resultlabel{GTE / KM-Prox} & \resultci{0.7615}{0.7570}{0.7659} & \resultci{0.7508}{0.7461}{0.7554} & \resultci{0.7301}{0.7253}{0.7346} & \resultci{0.7082}{0.7033}{0.7128} & \resultci{0.7287}{0.7246}{0.7329} \\[2pt]
\resultlabel{GTE / KM-Prox (validated)} & \resultci{0.7610}{0.7565}{0.7655} & \resultci{0.7510}{0.7463}{0.7555} & \resultci{0.7313}{0.7266}{0.7359} & \resultci{0.7080}{0.7032}{0.7126} & \resultci{0.7289}{0.7248}{0.7330} \\[2pt]
\resultlabel{GTE / MLP} & \resultci{0.7551}{0.7505}{0.7597} & \resultci{0.7462}{0.7417}{0.7508} & \resultci{0.7282}{0.7236}{0.7329} & \resultci{0.7021}{0.6975}{0.7067} & \resultci{0.7245}{0.7205}{0.7285} \\
\bottomrule
\end{tabular}
\end{table}

\begin{table}[H]
\setlength{\abovecaptionskip}{4pt}
\setlength{\belowcaptionskip}{6pt}
\centering\small
\setlength{\tabcolsep}{3pt}
\caption{RouterBench in-distribution utility with 95\% intervals, part 3 of 3. Mean averages all thirteen penalties.}
\label{tab:ci-routerbench-iid-3}
\begin{tabular}{@{}lrrrrr@{}}
\toprule
Method / $\lambda$ & 100 & 150 & 250 & 500 & Mean \\[2pt]
\midrule
\resultlabel{GTE / Avengers-Pro} & \resultci{0.6824}{0.6773}{0.6875} & \resultci{0.6620}{0.6571}{0.6672} & \resultci{0.6376}{0.6325}{0.6427} & \resultci{0.5794}{0.5740}{0.5850} & \resultci{0.7176}{0.7134}{0.7219} \\[2pt]
\resultlabel{GTE / Best single model} & \resultci{0.6443}{0.6388}{0.6498} & \resultci{0.6350}{0.6295}{0.6405} & \resultci{0.6164}{0.6108}{0.6220} & \resultci{0.5697}{0.5641}{0.5754} & \resultci{0.6994}{0.6951}{0.7035} \\[2pt]
\resultlabel{GTE / Family lookup} & \resultci{0.6753}{0.6704}{0.6802} & \resultci{0.6539}{0.6486}{0.6593} & \resultci{0.6292}{0.6238}{0.6348} & \resultci{0.5885}{0.5831}{0.5941} & \resultci{0.7212}{0.7172}{0.7252} \\[2pt]
\resultlabel{GTE / Family-cell lookup} & \resultci{0.6840}{0.6788}{0.6890} & \resultci{0.6598}{0.6547}{0.6650} & \resultci{0.6323}{0.6269}{0.6376} & \resultci{0.5936}{0.5879}{0.5992} & \resultci{0.7229}{0.7187}{0.7271} \\[2pt]
\resultlabel{GTE / KM-Prox} & \resultci{0.6915}{0.6868}{0.6965} & \resultci{0.6660}{0.6609}{0.6712} & \resultci{0.6399}{0.6346}{0.6452} & \resultci{0.6002}{0.5949}{0.6057} & \resultci{0.7287}{0.7246}{0.7329} \\[2pt]
\resultlabel{GTE / KM-Prox (validated)} & \resultci{0.6912}{0.6864}{0.6960} & \resultci{0.6659}{0.6610}{0.6708} & \resultci{0.6391}{0.6339}{0.6441} & \resultci{0.5994}{0.5941}{0.6050} & \resultci{0.7289}{0.7248}{0.7330} \\[2pt]
\resultlabel{GTE / MLP} & \resultci{0.6842}{0.6797}{0.6891} & \resultci{0.6623}{0.6575}{0.6671} & \resultci{0.6343}{0.6293}{0.6392} & \resultci{0.5953}{0.5904}{0.6006} & \resultci{0.7245}{0.7205}{0.7285} \\
\bottomrule
\end{tabular}
\end{table}

\begin{table}[H]
\setlength{\abovecaptionskip}{4pt}
\setlength{\belowcaptionskip}{6pt}
\centering\small
\setlength{\tabcolsep}{3pt}
\caption{RouterBench out-of-distribution utility with 95\% intervals, part 1 of 3. Mean averages all thirteen penalties.}
\label{tab:ci-routerbench-ood-1}
\begin{tabular}{@{}lrrrrrr@{}}
\toprule
Method / $\lambda$ & 0 & 1 & 2 & 5 & 10 & Mean \\[2pt]
\midrule
\resultlabel{GTE / Avengers-Pro} & \resultci{0.7905}{0.7871}{0.7942} & \resultci{0.7873}{0.7839}{0.7910} & \resultci{0.7841}{0.7807}{0.7878} & \resultci{0.7743}{0.7709}{0.7780} & \resultci{0.7581}{0.7547}{0.7618} & \resultci{0.6662}{0.6631}{0.6698} \\[2pt]
\resultlabel{GTE / Best single model} & \resultci{0.7921}{0.7887}{0.7958} & \resultci{0.7889}{0.7854}{0.7926} & \resultci{0.7856}{0.7822}{0.7893} & \resultci{0.7758}{0.7724}{0.7795} & \resultci{0.7596}{0.7561}{0.7632} & \resultci{0.6618}{0.6586}{0.6652} \\[2pt]
\resultlabel{GTE / KM-Prox} & \resultci{0.7921}{0.7887}{0.7958} & \resultci{0.7889}{0.7854}{0.7926} & \resultci{0.7856}{0.7822}{0.7893} & \resultci{0.7758}{0.7724}{0.7795} & \resultci{0.7595}{0.7560}{0.7632} & \resultci{0.6764}{0.6732}{0.6797} \\[2pt]
\resultlabel{GTE / KM-Prox (validated)} & \resultci{0.7921}{0.7887}{0.7958} & \resultci{0.7889}{0.7854}{0.7926} & \resultci{0.7856}{0.7822}{0.7893} & \resultci{0.7758}{0.7724}{0.7795} & \resultci{0.7576}{0.7542}{0.7613} & \resultci{0.6764}{0.6732}{0.6798} \\[2pt]
\resultlabel{GTE / MLP} & \resultci{0.7921}{0.7887}{0.7958} & \resultci{0.7889}{0.7854}{0.7926} & \resultci{0.7856}{0.7822}{0.7893} & \resultci{0.7758}{0.7724}{0.7795} & \resultci{0.7594}{0.7559}{0.7631} & \resultci{0.6755}{0.6722}{0.6789} \\[2pt]
\resultlabel{GTE / Realized oracle} & \resultci{0.9189}{0.9171}{0.9206} & \resultci{0.9186}{0.9169}{0.9203} & \resultci{0.9184}{0.9167}{0.9201} & \resultci{0.9177}{0.9160}{0.9194} & \resultci{0.9166}{0.9148}{0.9182} & \resultci{0.9030}{0.9012}{0.9047} \\[2pt]
\resultlabel{GTE / Ridge} & \resultci{0.7921}{0.7887}{0.7958} & \resultci{0.7889}{0.7854}{0.7926} & \resultci{0.7856}{0.7822}{0.7893} & \resultci{0.7758}{0.7724}{0.7795} & \resultci{0.7596}{0.7561}{0.7632} & \resultci{0.6796}{0.6765}{0.6829} \\
\bottomrule
\end{tabular}
\end{table}

\begin{table}[H]
\setlength{\abovecaptionskip}{4pt}
\setlength{\belowcaptionskip}{6pt}
\centering\small
\setlength{\tabcolsep}{3pt}
\caption{RouterBench out-of-distribution utility with 95\% intervals, part 2 of 3. Mean averages all thirteen penalties.}
\label{tab:ci-routerbench-ood-2}
\begin{tabular}{@{}lrrrrr@{}}
\toprule
Method / $\lambda$ & 20 & 30 & 50 & 75 & Mean \\[2pt]
\midrule
\resultlabel{GTE / Avengers-Pro} & \resultci{0.7250}{0.7216}{0.7288} & \resultci{0.6259}{0.6219}{0.6303} & \resultci{0.5724}{0.5684}{0.5769} & \resultci{0.5119}{0.5077}{0.5163} & \resultci{0.6662}{0.6631}{0.6698} \\[2pt]
\resultlabel{GTE / Best single model} & \resultci{0.7270}{0.7235}{0.7306} & \resultci{0.6484}{0.6445}{0.6526} & \resultci{0.5445}{0.5399}{0.5489} & \resultci{0.4959}{0.4911}{0.5002} & \resultci{0.6618}{0.6586}{0.6652} \\[2pt]
\resultlabel{GTE / KM-Prox} & \resultci{0.7267}{0.7232}{0.7304} & \resultci{0.6935}{0.6901}{0.6972} & \resultci{0.6267}{0.6233}{0.6305} & \resultci{0.5481}{0.5443}{0.5520} & \resultci{0.6764}{0.6732}{0.6797} \\[2pt]
\resultlabel{GTE / KM-Prox (validated)} & \resultci{0.7243}{0.7208}{0.7281} & \resultci{0.6923}{0.6888}{0.6961} & \resultci{0.6292}{0.6256}{0.6330} & \resultci{0.5527}{0.5489}{0.5567} & \resultci{0.6764}{0.6732}{0.6798} \\[2pt]
\resultlabel{GTE / MLP} & \resultci{0.7216}{0.7180}{0.7254} & \resultci{0.6778}{0.6741}{0.6816} & \resultci{0.5894}{0.5853}{0.5936} & \resultci{0.5258}{0.5212}{0.5302} & \resultci{0.6755}{0.6722}{0.6789} \\[2pt]
\resultlabel{GTE / Realized oracle} & \resultci{0.9142}{0.9125}{0.9159} & \resultci{0.9120}{0.9103}{0.9138} & \resultci{0.9080}{0.9063}{0.9098} & \resultci{0.9037}{0.9019}{0.9054} & \resultci{0.9030}{0.9012}{0.9047} \\[2pt]
\resultlabel{GTE / Ridge} & \resultci{0.7269}{0.7234}{0.7306} & \resultci{0.6907}{0.6873}{0.6944} & \resultci{0.5392}{0.5347}{0.5438} & \resultci{0.6054}{0.6009}{0.6099} & \resultci{0.6796}{0.6765}{0.6829} \\
\bottomrule
\end{tabular}
\end{table}

\begin{table}[H]
\setlength{\abovecaptionskip}{4pt}
\setlength{\belowcaptionskip}{6pt}
\centering\small
\setlength{\tabcolsep}{3pt}
\caption{RouterBench out-of-distribution utility with 95\% intervals, part 3 of 3. Mean averages all thirteen penalties.}
\label{tab:ci-routerbench-ood-3}
\begin{tabular}{@{}lrrrrr@{}}
\toprule
Method / $\lambda$ & 100 & 150 & 250 & 500 & Mean \\[2pt]
\midrule
\resultlabel{GTE / Avengers-Pro} & \resultci{0.6037}{0.5995}{0.6083} & \resultci{0.5994}{0.5948}{0.6041} & \resultci{0.5800}{0.5754}{0.5846} & \resultci{0.5483}{0.5437}{0.5529} & \resultci{0.6662}{0.6631}{0.6698} \\[2pt]
\resultlabel{GTE / Best single model} & \resultci{0.5983}{0.5936}{0.6028} & \resultci{0.5884}{0.5836}{0.5928} & \resultci{0.6167}{0.6124}{0.6212} & \resultci{0.4822}{0.4776}{0.4870} & \resultci{0.6618}{0.6586}{0.6652} \\[2pt]
\resultlabel{GTE / KM-Prox} & \resultci{0.4796}{0.4755}{0.4837} & \resultci{0.6351}{0.6308}{0.6397} & \resultci{0.6139}{0.6094}{0.6185} & \resultci{0.5669}{0.5624}{0.5715} & \resultci{0.6764}{0.6732}{0.6797} \\[2pt]
\resultlabel{GTE / KM-Prox (validated)} & \resultci{0.4747}{0.4706}{0.4790} & \resultci{0.6347}{0.6302}{0.6391} & \resultci{0.6163}{0.6119}{0.6208} & \resultci{0.5690}{0.5646}{0.5735} & \resultci{0.6764}{0.6732}{0.6798} \\[2pt]
\resultlabel{GTE / MLP} & \resultci{0.6171}{0.6126}{0.6218} & \resultci{0.6120}{0.6075}{0.6167} & \resultci{0.5921}{0.5876}{0.5966} & \resultci{0.5432}{0.5385}{0.5479} & \resultci{0.6755}{0.6722}{0.6789} \\[2pt]
\resultlabel{GTE / Realized oracle} & \resultci{0.8997}{0.8979}{0.9015} & \resultci{0.8921}{0.8903}{0.8939} & \resultci{0.8771}{0.8751}{0.8789} & \resultci{0.8414}{0.8393}{0.8435} & \resultci{0.9030}{0.9012}{0.9047} \\[2pt]
\resultlabel{GTE / Ridge} & \resultci{0.6336}{0.6292}{0.6382} & \resultci{0.6168}{0.6123}{0.6214} & \resultci{0.5994}{0.5949}{0.6040} & \resultci{0.5202}{0.5155}{0.5250} & \resultci{0.6796}{0.6765}{0.6829} \\
\bottomrule
\end{tabular}
\end{table}

\begin{table}[H]
\setlength{\abovecaptionskip}{4pt}
\setlength{\belowcaptionskip}{6pt}
\centering\small
\setlength{\tabcolsep}{3pt}
\caption{SPROUT in-distribution utility with 95\% intervals, part 1 of 3. Mean averages all thirteen penalties.}
\label{tab:ci-sprout-iid-1}
\begin{tabular}{@{}lrrrrrr@{}}
\toprule
Method / $\lambda$ & 0 & 0.5 & 1 & 2 & 5 & Mean \\[2pt]
\midrule
\resultlabel{GTE / Avengers-Pro} & \resultci{0.8648}{0.8612}{0.8683} & \resultci{0.8632}{0.8596}{0.8667} & \resultci{0.8618}{0.8582}{0.8654} & \resultci{0.8593}{0.8558}{0.8628} & \resultci{0.8536}{0.8500}{0.8572} & \resultci{0.8138}{0.8104}{0.8173} \\[2pt]
\resultlabel{GTE / Best single model} & \resultci{0.8434}{0.8395}{0.8471} & \resultci{0.8409}{0.8370}{0.8446} & \resultci{0.8385}{0.8346}{0.8422} & \resultci{0.8336}{0.8296}{0.8373} & \resultci{0.8188}{0.8148}{0.8227} & \resultci{0.7868}{0.7831}{0.7908} \\[2pt]
\resultlabel{GTE / Family lookup} & \resultci{0.8614}{0.8577}{0.8650} & \resultci{0.8596}{0.8559}{0.8632} & \resultci{0.8580}{0.8544}{0.8616} & \resultci{0.8546}{0.8509}{0.8583} & \resultci{0.8504}{0.8464}{0.8543} & \resultci{0.8097}{0.8061}{0.8132} \\[2pt]
\resultlabel{GTE / Family-cell lookup} & \resultci{0.8626}{0.8589}{0.8662} & \resultci{0.8605}{0.8568}{0.8641} & \resultci{0.8591}{0.8554}{0.8627} & \resultci{0.8553}{0.8515}{0.8591} & \resultci{0.8510}{0.8473}{0.8549} & \resultci{0.8122}{0.8088}{0.8158} \\[2pt]
\resultlabel{GTE / KM-Prox} & \resultci{0.8635}{0.8599}{0.8670} & \resultci{0.8620}{0.8585}{0.8655} & \resultci{0.8604}{0.8567}{0.8638} & \resultci{0.8584}{0.8549}{0.8618} & \resultci{0.8532}{0.8495}{0.8569} & \resultci{0.8148}{0.8114}{0.8183} \\[2pt]
\resultlabel{GTE / KM-Prox (validated)} & \resultci{0.8651}{0.8616}{0.8685} & \resultci{0.8642}{0.8606}{0.8676} & \resultci{0.8625}{0.8590}{0.8659} & \resultci{0.8611}{0.8575}{0.8646} & \resultci{0.8548}{0.8512}{0.8585} & \resultci{0.8166}{0.8132}{0.8200} \\[2pt]
\resultlabel{GTE / MLP} & \resultci{0.8577}{0.8543}{0.8611} & \resultci{0.8563}{0.8529}{0.8597} & \resultci{0.8563}{0.8528}{0.8597} & \resultci{0.8543}{0.8508}{0.8578} & \resultci{0.8477}{0.8441}{0.8512} & \resultci{0.8105}{0.8073}{0.8139} \\
\bottomrule
\end{tabular}
\end{table}

\begin{table}[H]
\setlength{\abovecaptionskip}{4pt}
\setlength{\belowcaptionskip}{6pt}
\centering\small
\setlength{\tabcolsep}{3pt}
\caption{SPROUT in-distribution utility with 95\% intervals, part 2 of 3. Mean averages all thirteen penalties.}
\label{tab:ci-sprout-iid-2}
\begin{tabular}{@{}lrrrrr@{}}
\toprule
Method / $\lambda$ & 10 & 25 & 50 & 75 & Mean \\[2pt]
\midrule
\resultlabel{GTE / Avengers-Pro} & \resultci{0.8446}{0.8410}{0.8482} & \resultci{0.8337}{0.8298}{0.8375} & \resultci{0.8188}{0.8150}{0.8228} & \resultci{0.8068}{0.8030}{0.8108} & \resultci{0.8138}{0.8104}{0.8173} \\[2pt]
\resultlabel{GTE / Best single model} & \resultci{0.8015}{0.7975}{0.8059} & \resultci{0.7973}{0.7929}{0.8018} & \resultci{0.7889}{0.7845}{0.7934} & \resultci{0.7804}{0.7760}{0.7850} & \resultci{0.7868}{0.7831}{0.7908} \\[2pt]
\resultlabel{GTE / Family lookup} & \resultci{0.8436}{0.8395}{0.8476} & \resultci{0.8269}{0.8229}{0.8310} & \resultci{0.8152}{0.8112}{0.8194} & \resultci{0.8037}{0.7997}{0.8079} & \resultci{0.8097}{0.8061}{0.8132} \\[2pt]
\resultlabel{GTE / Family-cell lookup} & \resultci{0.8448}{0.8411}{0.8488} & \resultci{0.8335}{0.8296}{0.8374} & \resultci{0.8205}{0.8165}{0.8245} & \resultci{0.8063}{0.8024}{0.8104} & \resultci{0.8122}{0.8088}{0.8158} \\[2pt]
\resultlabel{GTE / KM-Prox} & \resultci{0.8466}{0.8429}{0.8504} & \resultci{0.8342}{0.8304}{0.8382} & \resultci{0.8166}{0.8126}{0.8207} & \resultci{0.8056}{0.8016}{0.8097} & \resultci{0.8148}{0.8114}{0.8183} \\[2pt]
\resultlabel{GTE / KM-Prox (validated)} & \resultci{0.8477}{0.8440}{0.8515} & \resultci{0.8370}{0.8332}{0.8408} & \resultci{0.8204}{0.8165}{0.8243} & \resultci{0.8066}{0.8027}{0.8108} & \resultci{0.8166}{0.8132}{0.8200} \\[2pt]
\resultlabel{GTE / MLP} & \resultci{0.8381}{0.8345}{0.8417} & \resultci{0.8274}{0.8237}{0.8313} & \resultci{0.8138}{0.8102}{0.8175} & \resultci{0.8025}{0.7988}{0.8063} & \resultci{0.8105}{0.8073}{0.8139} \\
\bottomrule
\end{tabular}
\end{table}

\begin{table}[H]
\setlength{\abovecaptionskip}{4pt}
\setlength{\belowcaptionskip}{6pt}
\centering\small
\setlength{\tabcolsep}{3pt}
\caption{SPROUT in-distribution utility with 95\% intervals, part 3 of 3. Mean averages all thirteen penalties.}
\label{tab:ci-sprout-iid-3}
\begin{tabular}{@{}lrrrrr@{}}
\toprule
Method / $\lambda$ & 100 & 150 & 250 & 500 & Mean \\[2pt]
\midrule
\resultlabel{GTE / Avengers-Pro} & \resultci{0.7943}{0.7904}{0.7983} & \resultci{0.7731}{0.7693}{0.7773} & \resultci{0.7367}{0.7325}{0.7411} & \resultci{0.6689}{0.6643}{0.6738} & \resultci{0.8138}{0.8104}{0.8173} \\[2pt]
\resultlabel{GTE / Best single model} & \resultci{0.7720}{0.7676}{0.7766} & \resultci{0.7551}{0.7507}{0.7598} & \resultci{0.7214}{0.7168}{0.7262} & \resultci{0.6371}{0.6321}{0.6423} & \resultci{0.7868}{0.7831}{0.7908} \\[2pt]
\resultlabel{GTE / Family lookup} & \resultci{0.7927}{0.7886}{0.7970} & \resultci{0.7709}{0.7667}{0.7753} & \resultci{0.7313}{0.7270}{0.7358} & \resultci{0.6574}{0.6524}{0.6624} & \resultci{0.8097}{0.8061}{0.8132} \\[2pt]
\resultlabel{GTE / Family-cell lookup} & \resultci{0.7955}{0.7915}{0.7998} & \resultci{0.7752}{0.7711}{0.7795} & \resultci{0.7364}{0.7322}{0.7408} & \resultci{0.6584}{0.6536}{0.6635} & \resultci{0.8122}{0.8088}{0.8158} \\[2pt]
\resultlabel{GTE / KM-Prox} & \resultci{0.7948}{0.7908}{0.7989} & \resultci{0.7764}{0.7724}{0.7807} & \resultci{0.7434}{0.7391}{0.7478} & \resultci{0.6780}{0.6735}{0.6832} & \resultci{0.8148}{0.8114}{0.8183} \\[2pt]
\resultlabel{GTE / KM-Prox (validated)} & \resultci{0.7951}{0.7911}{0.7993} & \resultci{0.7766}{0.7726}{0.7809} & \resultci{0.7466}{0.7424}{0.7510} & \resultci{0.6777}{0.6731}{0.6826} & \resultci{0.8166}{0.8132}{0.8200} \\[2pt]
\resultlabel{GTE / MLP} & \resultci{0.7937}{0.7899}{0.7977} & \resultci{0.7746}{0.7706}{0.7787} & \resultci{0.7424}{0.7383}{0.7467} & \resultci{0.6719}{0.6675}{0.6767} & \resultci{0.8105}{0.8073}{0.8139} \\
\bottomrule
\end{tabular}
\end{table}

\begin{table}[H]
\setlength{\abovecaptionskip}{4pt}
\setlength{\belowcaptionskip}{6pt}
\centering\small
\setlength{\tabcolsep}{3pt}
\caption{SPROUT out-of-distribution utility with 95\% intervals, part 1 of 3. Mean averages all thirteen penalties.}
\label{tab:ci-sprout-ood-1}
\begin{tabular}{@{}lrrrrrr@{}}
\toprule
Method / $\lambda$ & 0 & 0.5 & 1 & 2 & 5 & Mean \\[2pt]
\midrule
\resultlabel{GTE / Avengers-Pro} & \resultci{0.7891}{0.7857}{0.7924} & \resultci{0.7883}{0.7849}{0.7916} & \resultci{0.7875}{0.7841}{0.7908} & \resultci{0.7874}{0.7840}{0.7907} & \resultci{0.7830}{0.7796}{0.7864} & \resultci{0.7439}{0.7408}{0.7470} \\[2pt]
\resultlabel{GTE / Best single model} & \resultci{0.8445}{0.8414}{0.8475} & \resultci{0.8420}{0.8389}{0.8451} & \resultci{0.8040}{0.8008}{0.8070} & \resultci{0.8010}{0.7978}{0.8041} & \resultci{0.7669}{0.7634}{0.7704} & \resultci{0.7575}{0.7544}{0.7604} \\[2pt]
\resultlabel{GTE / KM-Prox} & \resultci{0.7887}{0.7854}{0.7921} & \resultci{0.7873}{0.7840}{0.7906} & \resultci{0.7861}{0.7828}{0.7894} & \resultci{0.7837}{0.7803}{0.7869} & \resultci{0.7808}{0.7774}{0.7841} & \resultci{0.7460}{0.7429}{0.7491} \\[2pt]
\resultlabel{GTE / KM-Prox (validated)} & \resultci{0.7968}{0.7934}{0.8001} & \resultci{0.7948}{0.7914}{0.7981} & \resultci{0.7929}{0.7896}{0.7962} & \resultci{0.7889}{0.7855}{0.7921} & \resultci{0.7765}{0.7731}{0.7798} & \resultci{0.7483}{0.7453}{0.7513} \\[2pt]
\resultlabel{GTE / MLP} & \resultci{0.8088}{0.8055}{0.8121} & \resultci{0.8056}{0.8022}{0.8089} & \resultci{0.8125}{0.8092}{0.8158} & \resultci{0.8071}{0.8037}{0.8104} & \resultci{0.7971}{0.7936}{0.8005} & \resultci{0.7564}{0.7536}{0.7592} \\[2pt]
\resultlabel{GTE / Realized oracle} & \resultci{0.9766}{0.9753}{0.9778} & \resultci{0.9764}{0.9751}{0.9776} & \resultci{0.9762}{0.9750}{0.9774} & \resultci{0.9758}{0.9746}{0.9770} & \resultci{0.9747}{0.9734}{0.9759} & \resultci{0.9601}{0.9588}{0.9613} \\[2pt]
\resultlabel{GTE / Ridge} & \resultci{0.8328}{0.8296}{0.8359} & \resultci{0.8292}{0.8260}{0.8322} & \resultci{0.8256}{0.8225}{0.8288} & \resultci{0.8201}{0.8169}{0.8232} & \resultci{0.8097}{0.8065}{0.8129} & \resultci{0.7674}{0.7645}{0.7703} \\
\bottomrule
\end{tabular}
\end{table}

\begin{table}[H]
\setlength{\abovecaptionskip}{4pt}
\setlength{\belowcaptionskip}{6pt}
\centering\small
\setlength{\tabcolsep}{3pt}
\caption{SPROUT out-of-distribution utility with 95\% intervals, part 2 of 3. Mean averages all thirteen penalties.}
\label{tab:ci-sprout-ood-2}
\begin{tabular}{@{}lrrrrr@{}}
\toprule
Method / $\lambda$ & 10 & 25 & 50 & 75 & Mean \\[2pt]
\midrule
\resultlabel{GTE / Avengers-Pro} & \resultci{0.7822}{0.7789}{0.7856} & \resultci{0.7799}{0.7765}{0.7832} & \resultci{0.7663}{0.7629}{0.7698} & \resultci{0.7548}{0.7514}{0.7582} & \resultci{0.7439}{0.7408}{0.7470} \\[2pt]
\resultlabel{GTE / Best single model} & \resultci{0.7576}{0.7542}{0.7611} & \resultci{0.7641}{0.7607}{0.7676} & \resultci{0.7546}{0.7511}{0.7581} & \resultci{0.7451}{0.7416}{0.7486} & \resultci{0.7575}{0.7544}{0.7604} \\[2pt]
\resultlabel{GTE / KM-Prox} & \resultci{0.7796}{0.7762}{0.7829} & \resultci{0.7725}{0.7690}{0.7759} & \resultci{0.7579}{0.7543}{0.7614} & \resultci{0.7437}{0.7400}{0.7471} & \resultci{0.7460}{0.7429}{0.7491} \\[2pt]
\resultlabel{GTE / KM-Prox (validated)} & \resultci{0.7657}{0.7622}{0.7690} & \resultci{0.7699}{0.7665}{0.7732} & \resultci{0.7577}{0.7543}{0.7611} & \resultci{0.7443}{0.7409}{0.7478} & \resultci{0.7483}{0.7453}{0.7513} \\[2pt]
\resultlabel{GTE / MLP} & \resultci{0.7908}{0.7873}{0.7940} & \resultci{0.7744}{0.7712}{0.7776} & \resultci{0.7452}{0.7416}{0.7487} & \resultci{0.7354}{0.7318}{0.7389} & \resultci{0.7564}{0.7536}{0.7592} \\[2pt]
\resultlabel{GTE / Realized oracle} & \resultci{0.9730}{0.9717}{0.9742} & \resultci{0.9692}{0.9680}{0.9705} & \resultci{0.9642}{0.9629}{0.9655} & \resultci{0.9595}{0.9582}{0.9608} & \resultci{0.9601}{0.9588}{0.9613} \\[2pt]
\resultlabel{GTE / Ridge} & \resultci{0.7895}{0.7862}{0.7927} & \resultci{0.7667}{0.7633}{0.7701} & \resultci{0.7527}{0.7492}{0.7562} & \resultci{0.7430}{0.7395}{0.7466} & \resultci{0.7674}{0.7645}{0.7703} \\
\bottomrule
\end{tabular}
\end{table}

\begin{table}[H]
\setlength{\abovecaptionskip}{4pt}
\setlength{\belowcaptionskip}{6pt}
\centering\small
\setlength{\tabcolsep}{3pt}
\caption{SPROUT out-of-distribution utility with 95\% intervals, part 3 of 3. Mean averages all thirteen penalties.}
\label{tab:ci-sprout-ood-3}
\begin{tabular}{@{}lrrrrr@{}}
\toprule
Method / $\lambda$ & 100 & 150 & 250 & 500 & Mean \\[2pt]
\midrule
\resultlabel{GTE / Avengers-Pro} & \resultci{0.7423}{0.7390}{0.7458} & \resultci{0.7175}{0.7141}{0.7211} & \resultci{0.6413}{0.6375}{0.6449} & \resultci{0.5510}{0.5469}{0.5550} & \resultci{0.7439}{0.7408}{0.7470} \\[2pt]
\resultlabel{GTE / Best single model} & \resultci{0.7356}{0.7321}{0.7392} & \resultci{0.7166}{0.7131}{0.7203} & \resultci{0.7218}{0.7181}{0.7255} & \resultci{0.5930}{0.5890}{0.5971} & \resultci{0.7575}{0.7544}{0.7604} \\[2pt]
\resultlabel{GTE / KM-Prox} & \resultci{0.7319}{0.7282}{0.7354} & \resultci{0.7123}{0.7086}{0.7159} & \resultci{0.6822}{0.6785}{0.6859} & \resultci{0.5908}{0.5866}{0.5951} & \resultci{0.7460}{0.7429}{0.7491} \\[2pt]
\resultlabel{GTE / KM-Prox (validated)} & \resultci{0.7330}{0.7295}{0.7364} & \resultci{0.7215}{0.7180}{0.7251} & \resultci{0.6907}{0.6870}{0.6944} & \resultci{0.5951}{0.5909}{0.5990} & \resultci{0.7483}{0.7453}{0.7513} \\[2pt]
\resultlabel{GTE / MLP} & \resultci{0.7313}{0.7277}{0.7348} & \resultci{0.7164}{0.7129}{0.7201} & \resultci{0.7218}{0.7181}{0.7255} & \resultci{0.5863}{0.5821}{0.5905} & \resultci{0.7564}{0.7536}{0.7592} \\[2pt]
\resultlabel{GTE / Realized oracle} & \resultci{0.9549}{0.9535}{0.9562} & \resultci{0.9463}{0.9449}{0.9477} & \resultci{0.9316}{0.9300}{0.9330} & \resultci{0.9026}{0.9008}{0.9042} & \resultci{0.9601}{0.9588}{0.9613} \\[2pt]
\resultlabel{GTE / Ridge} & \resultci{0.7339}{0.7304}{0.7375} & \resultci{0.7157}{0.7121}{0.7193} & \resultci{0.7210}{0.7172}{0.7246} & \resultci{0.6364}{0.6322}{0.6404} & \resultci{0.7674}{0.7645}{0.7703} \\
\bottomrule
\end{tabular}
\end{table}

\clearpage
\subsection{SharedTrunkNet results and robustness}
\label{app:sharedtrunknet-results}

The implementation and selection procedures are specified in Appendix~\ref{app:sharedtrunknet}.
Table~\ref{tab:nvidia-mean-ci} also reports the fixed-L21 and common-layer controls.
The per-model selection procedure used in the main comparison has the highest mean among these three SharedTrunkNet variants on both benchmarks.

\begin{table}[H]
\setlength{\abovecaptionskip}{4pt}
\setlength{\belowcaptionskip}{6pt}
\centering\small
\setlength{\tabcolsep}{6pt}
\caption{Out-of-distribution mean utility over thirteen penalties, with 95\% intervals.}
\label{tab:nvidia-mean-ci}
\begin{tabular}{@{}lrr@{}}
\toprule
Method & RouterBench & SPROUT \\
\midrule

\resultlabel{SharedTrunkNet: fixed L21} & \resultci{0.6841}{0.6809}{0.6875} & \resultci{0.7812}{0.7783}{0.7841} \\[2pt]

\resultlabel{SharedTrunkNet: common selected layer} & \resultci{0.6803}{0.6771}{0.6837} & \resultci{0.7806}{0.7777}{0.7835} \\[2pt]

\resultlabel{SharedTrunkNet: per-model selected layers} & \resultci{0.6949}{0.6918}{0.6983} & \resultci{0.7829}{0.7800}{0.7857} \\[2pt]

\midrule

\resultlabel{L21 kNN-100} & \resultci{0.7007}{0.6972}{0.7043} & \resultci{0.7731}{0.7701}{0.7760} \\[2pt]

\resultlabel{L21 Family Adaptation} & \resultci{0.7157}{0.7125}{0.7191} & \resultci{0.7767}{0.7737}{0.7797} \\[2pt]

\bottomrule
\end{tabular}
\end{table}

\paragraph{Paired utility differences.}
Table~\ref{tab:sharedtrunknet-differences} subtracts the main SharedTrunkNet variant from each L21 method, using the same queries and averaging all thirteen penalties.
Intervals use the paired query bootstrap in Appendix~\ref{app:routing-ci}.
Both L21 methods have higher mean utility on RouterBench and lower mean utility on SPROUT; none of these four difference intervals includes zero.
These intervals condition on the fitted routers and evaluated families.

\begin{table}[H]
\setlength{\abovecaptionskip}{4pt}
\setlength{\belowcaptionskip}{6pt}
\centering\small
\setlength{\tabcolsep}{6pt}
\caption{Mean utility difference from SharedTrunkNet, with paired 95\% intervals. Positive values favor the L21 method.}
\label{tab:sharedtrunknet-differences}
\begin{tabular}{@{}lrr@{}}
\toprule
Comparison & RouterBench & SPROUT \\
\midrule

\resultlabel{L21 kNN-100 minus SharedTrunkNet} & \resultci{$+$0.0058}{$+$0.0043}{$+$0.0072} & \resultci{\tableminus 0.0097}{\tableminus 0.0112}{\tableminus 0.0083} \\[2pt]

\resultlabel{L21 Family Adaptation minus SharedTrunkNet} & \resultci{$+$0.0208}{$+$0.0195}{$+$0.0221} & \resultci{\tableminus 0.0062}{\tableminus 0.0073}{\tableminus 0.0050} \\[2pt]

\bottomrule
\end{tabular}
\end{table}

\paragraph{Results at every penalty.}
Tables~\ref{tab:sharedtrunknet-routerbench} and~\ref{tab:sharedtrunknet-sprout} report utility at all thirteen cost penalties.
SharedTrunkNet denotes the per-model selection procedure used in the main text.
On RouterBench, the Family Adaptation has higher utility at 13 of 13 penalties and kNN-100 at 7 of 13.
On SPROUT, each has higher utility at 3 of 13 penalties.
These counts describe the observed comparisons; penalties are not independent statistical replications.

\begin{table}[H]
\setlength{\abovecaptionskip}{4pt}
\setlength{\belowcaptionskip}{6pt}
\centering\small
\setlength{\tabcolsep}{6pt}
\caption{RouterBench: out-of-distribution utility at every penalty, with 95\% intervals.}
\label{tab:sharedtrunknet-routerbench}
\begin{tabular}{@{}rrrr@{}}
\toprule
$\lambda$ & SharedTrunkNet & L21 kNN-100 & L21 Family Adaptation \\
\midrule

\resultlabel{0} & \resultci{0.7916}{0.7882}{0.7952} & \resultci{0.7854}{0.7818}{0.7891} & \resultci{0.7921}{0.7887}{0.7958} \\[2pt]

\resultlabel{1} & \resultci{0.7887}{0.7853}{0.7925} & \resultci{0.7805}{0.7769}{0.7843} & \resultci{0.7889}{0.7854}{0.7926} \\[2pt]

\resultlabel{2} & \resultci{0.7855}{0.7820}{0.7891} & \resultci{0.7771}{0.7734}{0.7809} & \resultci{0.7856}{0.7822}{0.7893} \\[2pt]

\resultlabel{5} & \resultci{0.7754}{0.7719}{0.7790} & \resultci{0.7678}{0.7642}{0.7716} & \resultci{0.7759}{0.7724}{0.7796} \\[2pt]

\resultlabel{10} & \resultci{0.7585}{0.7551}{0.7622} & \resultci{0.7587}{0.7549}{0.7626} & \resultci{0.7661}{0.7624}{0.7698} \\[2pt]

\resultlabel{20} & \resultci{0.7207}{0.7171}{0.7246} & \resultci{0.7439}{0.7401}{0.7478} & \resultci{0.7523}{0.7486}{0.7560} \\[2pt]

\resultlabel{30} & \resultci{0.6801}{0.6764}{0.6841} & \resultci{0.7279}{0.7240}{0.7319} & \resultci{0.7385}{0.7348}{0.7423} \\[2pt]

\resultlabel{50} & \resultci{0.6786}{0.6746}{0.6826} & \resultci{0.6994}{0.6954}{0.7037} & \resultci{0.7126}{0.7090}{0.7164} \\[2pt]

\resultlabel{75} & \resultci{0.6780}{0.6739}{0.6823} & \resultci{0.6691}{0.6648}{0.6735} & \resultci{0.6966}{0.6927}{0.7005} \\[2pt]

\resultlabel{100} & \resultci{0.6323}{0.6279}{0.6366} & \resultci{0.6462}{0.6418}{0.6508} & \resultci{0.6689}{0.6647}{0.6733} \\[2pt]

\resultlabel{150} & \resultci{0.6078}{0.6034}{0.6125} & \resultci{0.6189}{0.6143}{0.6235} & \resultci{0.6350}{0.6307}{0.6395} \\[2pt]

\resultlabel{250} & \resultci{0.6015}{0.5972}{0.6060} & \resultci{0.5899}{0.5852}{0.5946} & \resultci{0.6191}{0.6147}{0.6235} \\[2pt]

\resultlabel{500} & \resultci{0.5352}{0.5307}{0.5400} & \resultci{0.5441}{0.5394}{0.5487} & \resultci{0.5726}{0.5683}{0.5771} \\[2pt]

\midrule

\resultlabel{Mean} & \resultci{0.6949}{0.6918}{0.6983} & \resultci{0.7007}{0.6972}{0.7043} & \resultci{0.7157}{0.7125}{0.7191} \\[2pt]

\bottomrule
\end{tabular}
\end{table}

\begin{table}[H]
\setlength{\abovecaptionskip}{4pt}
\setlength{\belowcaptionskip}{6pt}
\centering\small
\setlength{\tabcolsep}{6pt}
\caption{SPROUT: out-of-distribution utility at every penalty, with 95\% intervals.}
\label{tab:sharedtrunknet-sprout}
\begin{tabular}{@{}rrrr@{}}
\toprule
$\lambda$ & SharedTrunkNet & L21 kNN-100 & L21 Family Adaptation \\
\midrule

\resultlabel{0} & \resultci{0.8212}{0.8179}{0.8244} & \resultci{0.8243}{0.8212}{0.8275} & \resultci{0.8163}{0.8130}{0.8196} \\[2pt]

\resultlabel{0.5} & \resultci{0.8321}{0.8291}{0.8354} & \resultci{0.8225}{0.8192}{0.8257} & \resultci{0.8129}{0.8095}{0.8161} \\[2pt]

\resultlabel{1} & \resultci{0.8145}{0.8113}{0.8178} & \resultci{0.8200}{0.8169}{0.8232} & \resultci{0.8094}{0.8061}{0.8126} \\[2pt]

\resultlabel{2} & \resultci{0.8091}{0.8059}{0.8124} & \resultci{0.8170}{0.8137}{0.8202} & \resultci{0.8025}{0.7992}{0.8058} \\[2pt]

\resultlabel{5} & \resultci{0.8118}{0.8085}{0.8150} & \resultci{0.8073}{0.8040}{0.8106} & \resultci{0.8070}{0.8035}{0.8103} \\[2pt]

\resultlabel{10} & \resultci{0.8155}{0.8122}{0.8188} & \resultci{0.8009}{0.7974}{0.8042} & \resultci{0.8085}{0.8050}{0.8117} \\[2pt]

\resultlabel{25} & \resultci{0.8074}{0.8042}{0.8106} & \resultci{0.7946}{0.7911}{0.7979} & \resultci{0.8042}{0.8008}{0.8075} \\[2pt]

\resultlabel{50} & \resultci{0.7972}{0.7939}{0.8004} & \resultci{0.7830}{0.7796}{0.7864} & \resultci{0.7957}{0.7924}{0.7991} \\[2pt]

\resultlabel{75} & \resultci{0.7746}{0.7712}{0.7779} & \resultci{0.7716}{0.7682}{0.7751} & \resultci{0.7873}{0.7839}{0.7906} \\[2pt]

\resultlabel{100} & \resultci{0.7847}{0.7813}{0.7881} & \resultci{0.7606}{0.7571}{0.7641} & \resultci{0.7772}{0.7739}{0.7806} \\[2pt]

\resultlabel{150} & \resultci{0.7566}{0.7531}{0.7601} & \resultci{0.7384}{0.7348}{0.7419} & \resultci{0.7571}{0.7536}{0.7605} \\[2pt]

\resultlabel{250} & \resultci{0.7134}{0.7096}{0.7172} & \resultci{0.6986}{0.6947}{0.7023} & \resultci{0.7210}{0.7173}{0.7247} \\[2pt]

\resultlabel{500} & \resultci{0.6392}{0.6351}{0.6431} & \resultci{0.6119}{0.6079}{0.6160} & \resultci{0.5975}{0.5933}{0.6016} \\[2pt]

\midrule

\resultlabel{Mean} & \resultci{0.7829}{0.7800}{0.7857} & \resultci{0.7731}{0.7701}{0.7760} & \resultci{0.7767}{0.7737}{0.7797} \\[2pt]

\bottomrule
\end{tabular}
\end{table}

\clearpage\subsection{Breaking Down the Routing Gain by Family}
\label{app:routing-gain-by-family}
We compare kNN-100 with GTE and Qwen L21 embeddings on each held-out family. Each query is routed using only training queries from the other families, with the same candidate models and cost penalties for both representations. We average realized utility across the thirteen penalties and across queries within each family. The gain $\Delta$ is L21 minus GTE.

\paragraph{Equal query and equal family weights.}
Let $\bar u_j$ be the mean utility within family $j$, containing $n_j$ queries, and let $N=\sum_{j=1}^{J}n_j$. We report both
\[
\bar u_{\mathrm{query}}=\sum_{j=1}^{J}\frac{n_j}{N}\bar u_j,
\qquad
\bar u_{\mathrm{family}}=\frac{1}{J}\sum_{j=1}^{J}\bar u_j.
\]
The equal-query average gives every test query the same weight, so larger families contribute more. It matches the aggregation used in the main text. The equal-family average first normalizes each family by its query count and then gives every family the same weight: $1/7$ on RouterBench and $1/6$ on SPROUT. Both summaries use the same saved routing decisions; the individual family rows are identical under either aggregation.

\paragraph{Confidence intervals.}
We use the same 2,000 paired bootstrap samples as Appendix~\ref{app:routing-ci}, resampling queries within each held-out family and retaining both representations and all penalties together. Each average and paired gain is recomputed in every sample. Brackets report the 2.5th and 97.5th percentiles. These pointwise intervals condition on the evaluated families, splits and fitted routers.

\begin{table}[H]
\centering
\small
\setlength{\tabcolsep}{8pt}
\caption{RouterBench: kNN-100 utility by held-out family and under both averaging schemes, with 95\% intervals. $\Delta$ is L21 minus GTE.}
\label{tab:family-routing-routerbench}
\begin{tabular}{@{}lrrrr@{}}
\toprule
Family / aggregation & Queries & GTE & L21 & $\Delta$ \\[2pt]
\midrule
ARC-Challenge & 1,470 & \resultci{0.9080}{0.8979}{0.9177} & \resultci{0.9126}{0.9015}{0.9232} & \resultci{+0.0046}{-0.0021}{0.0107} \\[3pt]
GSM8K & 7,450 & \resultci{0.4125}{0.4088}{0.4162} & \resultci{0.5723}{0.5690}{0.5755} & \resultci{+0.1598}{0.1567}{0.1628} \\[3pt]
HellaSwag & 10,042 & \resultci{0.7220}{0.7155}{0.7288} & \resultci{0.7263}{0.7194}{0.7330} & \resultci{+0.0043}{0.0006}{0.0081} \\[3pt]
MBPP & 425 & \resultci{0.5316}{0.4922}{0.5700} & \resultci{0.5730}{0.5323}{0.6112} & \resultci{+0.0414}{0.0123}{0.0703} \\[3pt]
MMLU & 14,037 & \resultci{0.7253}{0.7192}{0.7316} & \resultci{0.7287}{0.7223}{0.7355} & \resultci{+0.0034}{0.0002}{0.0066} \\[3pt]
MT-Bench & 80 & \resultci{0.3995}{0.3170}{0.4805} & \resultci{0.5922}{0.5212}{0.6589} & \resultci{+0.1927}{0.1155}{0.2836} \\[3pt]
WinoGrande & 1,267 & \resultci{0.7151}{0.6963}{0.7343} & \resultci{0.7462}{0.7279}{0.7655} & \resultci{+0.0311}{0.0193}{0.0444} \\[3pt]
\midrule
Equal-query average & 34,771 & \resultci{0.6615}{0.6583}{0.6651} & \resultci{0.7007}{0.6972}{0.7043} & \resultci{+0.0391}{0.0371}{0.0410} \\[3pt]
Equal-family average & 34,771 & \resultci{0.6306}{0.6169}{0.6439} & \resultci{0.6930}{0.6810}{0.7046} & \resultci{+0.0625}{0.0505}{0.0756} \\[3pt]
\bottomrule
\end{tabular}
\end{table}

Changing the representation increases mean utility in all seven RouterBench families, with paired 95\% confidence intervals excluding zero in six. Most of the query-weighted gain comes from GSM8K, where utility increases from 0.4125 to 0.5723. This substantial improvement occurs in one of the larger families where GTE performs poorly, while gains also extend to the other families. MT-Bench, where GTE performs worst, shows the largest per-family improvement (+0.1927), although its contribution to the query-weighted average is small because it contains only 80 queries.

The particularly large gains on GSM8K and MT-Bench may not be coincidental. In this representation, GSM8K retrieves neighbors mainly from MBPP and MT-Bench instead of MMLU, while MT-Bench retrieves neighbors mainly from GSM8K and MBPP instead of MMLU and HellaSwag. For predicting relative model rewards, MT-Bench is the best-ranked training family for GSM8K, while GSM8K and MBPP rank first and third for MT-Bench, compared with sixth and fifth for MMLU and HellaSwag, respectively. These shifts provide a plausible explanation for the large gains.

\begin{table}[H]
\centering
\small
\setlength{\tabcolsep}{8pt}
\caption{SPROUT: kNN-100 utility by held-out family and under both averaging schemes, with 95\% intervals. $\Delta$ is L21 minus GTE.}
\label{tab:family-routing-sprout}
\begin{tabular}{@{}lrrrr@{}}
\toprule
Family / aggregation & Queries & GTE & L21 & $\Delta$ \\[2pt]
\midrule
GPQA & 540 & \resultci{0.4037}{0.3692}{0.4384} & \resultci{0.4715}{0.4399}{0.5038} & \resultci{+0.0678}{0.0375}{0.0989} \\[3pt]
MATH & 9,884 & \resultci{0.7601}{0.7526}{0.7675} & \resultci{0.7635}{0.7561}{0.7709} & \resultci{+0.0034}{0.0003}{0.0067} \\[3pt]
MMLU-Pro & 11,786 & \resultci{0.6698}{0.6628}{0.6772} & \resultci{0.6696}{0.6626}{0.6767} & \resultci{-0.0002}{-0.0041}{0.0035} \\[3pt]
MuSR & 748 & \resultci{0.5500}{0.5216}{0.5785} & \resultci{0.5149}{0.4840}{0.5460} & \resultci{-0.0351}{-0.0550}{-0.0153} \\[3pt]
OpenHermes & 19,340 & \resultci{0.8092}{0.8060}{0.8123} & \resultci{0.8505}{0.8472}{0.8538} & \resultci{+0.0414}{0.0391}{0.0437} \\[3pt]
RAGBench & 1,589 & \resultci{0.8539}{0.8435}{0.8646} & \resultci{0.8831}{0.8732}{0.8926} & \resultci{+0.0292}{0.0210}{0.0370} \\[3pt]
\midrule
Equal-query average & 43,887 & \resultci{0.7529}{0.7498}{0.7559} & \resultci{0.7731}{0.7701}{0.7760} & \resultci{+0.0202}{0.0185}{0.0220} \\[3pt]
Equal-family average & 43,887 & \resultci{0.6744}{0.6664}{0.6823} & \resultci{0.6922}{0.6845}{0.7001} & \resultci{+0.0177}{0.0113}{0.0238} \\[3pt]
\bottomrule
\end{tabular}
\end{table}

On SPROUT, significantly positive gains are spread across four families, while MMLU-Pro remains approximately unchanged. OpenHermes contributes most to the query-weighted improvement. MuSR is the only family with a significantly negative gain; the loss on this family could be further investigated.

In summary, mean utility increases in 11 of the 13 families. Paired 95\% confidence intervals exclude zero for ten gains and one decrease, on MuSR; the differences on ARC-Challenge and MMLU-Pro remain compatible with zero.

\clearpage

\section{Layer choice and robustness across depth}\label{app:layer-sweep}
\paragraph{Origin of the fixed layer.}
L21 was fixed following exploratory ID difficulty prediction on an earlier RouterBench panel. A ridge probe compared layers $\{0,7,14,21,28\}$ with last-token and mean pooling, predicting the number of models solving each query. L21 with last-token pooling gave the largest reduction in squared error beyond a family-mean predictor. This historical choice used RouterBench outcomes. We therefore test robustness across layers.

\paragraph{Matched routing sweep.}
We test whether the improvement depends on choosing L21 by repeating held-out-family routing with every upper-half layer, $\ell=15,\ldots,28$, of Qwen2.5-7B-Instruct. Each representation is the unit-normalized state of the last input token, \texttt{hidden\_states[$\ell$]}, using the same raw prompts. Queries, candidate models, splits, cost penalties and routing rules are fixed. The L21 results are the same results reported in the main text.

\begin{table}[H]
\setlength{\abovecaptionskip}{4pt}
\setlength{\belowcaptionskip}{6pt}
\centering\small
\setlength{\tabcolsep}{6pt}
\caption{Held-out-family mean utility over all thirteen penalties.}
\label{tab:layer-sweep}
\begin{tabular}{@{}lrrrrrr@{}}
\toprule
& \multicolumn{3}{c}{RouterBench} & \multicolumn{3}{c}{SPROUT} \\
\cmidrule(lr){2-4}\cmidrule(lr){5-7}
Layer & kNN-100 & Validated kNN & Family Adaptation & kNN-100 & Validated kNN & Family Adaptation \\
\midrule
GTE & 0.6615 & 0.6566 & 0.7006 & 0.7529 & 0.7534 & 0.7603 \\
\midrule
15 & 0.6933 & 0.6895 & 0.7157 & 0.7671 & 0.7644 & 0.7592 \\
16 & 0.6865 & 0.6878 & 0.7153 & 0.7667 & 0.7647 & 0.7592 \\
17 & 0.6890 & 0.6897 & 0.7165 & 0.7665 & 0.7645 & 0.7593 \\
18 & 0.7013 & 0.6977 & 0.7170 & 0.7673 & 0.7645 & 0.7610 \\
19 & 0.7038 & 0.7061 & 0.7167 & 0.7641 & 0.7600 & 0.7612 \\
20 & 0.7032 & 0.7051 & 0.7165 & 0.7719 & 0.7685 & 0.7751 \\
21 & 0.7007 & 0.7034 & 0.7157 & 0.7731 & 0.7739 & 0.7767 \\
22 & 0.7011 & 0.7034 & 0.7169 & 0.7754 & 0.7775 & 0.7767 \\
23 & 0.7037 & 0.7047 & 0.7165 & 0.7758 & 0.7777 & 0.7751 \\
24 & 0.7024 & 0.7034 & 0.7146 & 0.7762 & 0.7777 & 0.7758 \\
25 & 0.6959 & 0.6969 & 0.7148 & 0.7773 & 0.7784 & 0.7769 \\
26 & 0.6973 & 0.6962 & 0.7136 & 0.7765 & 0.7773 & 0.7769 \\
27 & 0.6963 & 0.6954 & 0.7140 & 0.7745 & 0.7774 & 0.7798 \\
28 & 0.6901 & 0.6782 & 0.7123 & 0.7636 & 0.7670 & 0.7726 \\
\bottomrule
\end{tabular}
\end{table}

Both kNN rules outperform their GTE counterparts at every tested layer on both benchmarks. The result is therefore not specific to L21. Layer choice still affects performance: SPROUT's Family Adaptation falls slightly below GTE at layers 15--17 and exceeds it at layers 18--28.

As a small additional check, we repeat in-distribution routing at L23 and L25 (Table~\ref{tab:layer-iid}). These two layers were chosen arbitrarily from the tested range, without using performance results to guide the choice. Their mean utilities remain within 0.0006 of L21 for each router and benchmark.
\begin{table}[H]
\setlength{\abovecaptionskip}{4pt}
\setlength{\belowcaptionskip}{6pt}
\centering\small
\setlength{\tabcolsep}{6pt}
\caption{ID mean utility over five splits and thirteen penalties.}
\label{tab:layer-iid}
\begin{tabular}{@{}lrrrrrr@{}}
\toprule
& \multicolumn{3}{c}{RouterBench} & \multicolumn{3}{c}{SPROUT} \\
\cmidrule(lr){2-4}\cmidrule(lr){5-7}
Layer & kNN-100 & Validated kNN & Family Adaptation & kNN-100 & Validated kNN & Family Adaptation \\
\midrule
21 & 0.7352 & 0.7360 & 0.7217 & 0.8206 & 0.8212 & 0.8096 \\
23 & 0.7356 & 0.7366 & 0.7218 & 0.8202 & 0.8209 & 0.8097 \\
25 & 0.7354 & 0.7355 & 0.7219 & 0.8203 & 0.8211 & 0.8096 \\
\bottomrule
\end{tabular}
\end{table}

\clearpage
\section{Cost, Reward and Answer-Format Controls}
\label{app:cost-reward-format}

\paragraph{Predictions and evaluation.}
We use the same out-of-distribution splits as the main comparison: each of the seven RouterBench and six SPROUT families is held out in turn.
All 34,771 RouterBench and 43,887 SPROUT queries are evaluated once.
For each test query $x$ and candidate model $m$, kNN-100 predicts reward $\widehat r(x,m)$ and cost $\widehat c(x,m)$ by averaging that model's recorded outcomes over the 100 nearest training queries.
For each encoder, neighbors are retrieved by cosine similarity.
Cost is the total dollar cost used in the routing objective, including both input and output costs where available; we do not measure answer length separately.

\paragraph{Uncertainty.}
All 95\% intervals use 2,000 paired bootstrap samples of queries within each held-out family, keeping family sizes fixed and using the same sampled queries across methods and penalties (Appendix~\ref{app:routing-ci}).
Each bootstrap mean is computed over the full penalty grid before taking interval endpoints.
These intervals condition on the evaluated families, frozen training pools and predictors.

\subsection{Prediction errors}\label{app:prediction-errors}
Mean squared error (MSE) and mean absolute error (MAE) average equally over queries and candidate models.
To examine relative model preferences, we also center predictions and recorded outcomes separately across models within each query before computing the same errors.
Centering removes a common increase or decrease shared by all models.
The training-mean reference predicts each model's average reward and cost over all available training queries.
Table~\ref{tab:cost-reward-errors} reports the errors; Table~\ref{tab:cost-reward-relative} reports $100(1-E_{\mathrm{L21}}/E_{\mathrm{GTE}})$, where $E$ is the corresponding error.
Positive percentages mean that L21 has lower error.
Errors for reward and dollar cost have different units and should not be compared directly.

\begin{table}[H]
\setlength{\abovecaptionskip}{4pt}
\setlength{\belowcaptionskip}{6pt}
\centering\small
\setlength{\tabcolsep}{5pt}
\caption{Reward and cost prediction errors, with 95\% intervals. Cost MSE and MAE are multiplied by $10^6$ and $10^3$, respectively, for readability.}
\label{tab:cost-reward-errors}
\begin{tabular}{@{}llrrrr@{}}
\toprule
Benchmark & Predictor & Reward MSE & Reward MAE & Cost MSE & Cost MAE \\
\midrule
\resultlabel{RouterBench} & \resultlabel{GTE} & \resultci{0.2061}{0.2053}{0.2070} & \resultci{0.4077}{0.4068}{0.4086} & \resultci{2.5186}{2.4758}{2.5611} & \resultci{0.5757}{0.5725}{0.5788} \\[2pt]
\resultlabel{RouterBench} & \resultlabel{L21} & \resultci{0.1780}{0.1770}{0.1791} & \resultci{0.3383}{0.3370}{0.3395} & \resultci{1.0323}{1.0077}{1.0580} & \resultci{0.3185}{0.3160}{0.3208} \\[2pt]
\resultlabel{RouterBench} & \resultlabel{GTE, matching format} & \resultci{0.1902}{0.1894}{0.1909} & \resultci{0.3852}{0.3844}{0.3860} & \resultci{0.8648}{0.8402}{0.8895} & \resultci{0.3406}{0.3382}{0.3428} \\[2pt]
\resultlabel{RouterBench} & \resultlabel{Training-model means} & \resultci{0.1899}{0.1893}{0.1905} & \resultci{0.3943}{0.3937}{0.3949} & \resultci{2.8637}{2.8255}{2.9016} & \resultci{0.7833}{0.7806}{0.7863} \\[2pt]
\midrule
\resultlabel{SPROUT} & \resultlabel{GTE} & \resultci{0.2073}{0.2064}{0.2082} & \resultci{0.4031}{0.4020}{0.4042} & \resultci{4.8932}{4.5794}{5.2115} & \resultci{0.7142}{0.7086}{0.7199} \\[2pt]
\resultlabel{SPROUT} & \resultlabel{L21} & \resultci{0.1614}{0.1604}{0.1624} & \resultci{0.3221}{0.3209}{0.3234} & \resultci{5.0209}{4.7157}{5.3337} & \resultci{0.7436}{0.7377}{0.7493} \\[2pt]
\resultlabel{SPROUT} & \resultlabel{GTE, matching format} & \resultci{0.1773}{0.1763}{0.1783} & \resultci{0.3478}{0.3467}{0.3489} & \resultci{5.2243}{4.9110}{5.5502} & \resultci{0.7163}{0.7106}{0.7222} \\[2pt]
\resultlabel{SPROUT} & \resultlabel{Training-model means} & \resultci{0.1921}{0.1912}{0.1930} & \resultci{0.3787}{0.3778}{0.3798} & \resultci{5.6598}{5.3400}{5.9843} & \resultci{0.7527}{0.7467}{0.7586} \\[2pt]
\bottomrule
\end{tabular}
\end{table}

\begin{table}[H]
\setlength{\abovecaptionskip}{4pt}
\setlength{\belowcaptionskip}{6pt}
\centering\small
\setlength{\tabcolsep}{5pt}
\caption{L21 error reduction relative to GTE (\%), with paired 95\% intervals. Negative values indicate higher error.}
\label{tab:cost-reward-relative}
\begin{tabular}{@{}lllrr@{}}
\toprule
Target & Error & Centering & RouterBench & SPROUT \\
\midrule
\resultlabel{Reward} & \resultlabel{MSE} & \resultlabel{None} & \resultci{$+$13.64}{$+$13.23}{$+$14.05} & \resultci{$+$22.15}{$+$21.67}{$+$22.64} \\[2pt]
\resultlabel{Reward} & \resultlabel{MAE} & \resultlabel{None} & \resultci{$+$17.03}{$+$16.79}{$+$17.27} & \resultci{$+$20.07}{$+$19.74}{$+$20.41} \\[2pt]
\resultlabel{Reward} & \resultlabel{MSE} & \resultlabel{Across models} & \resultci{$+$5.84}{$+$5.58}{$+$6.10} & \resultci{$+$2.43}{$+$2.26}{$+$2.59} \\[2pt]
\resultlabel{Reward} & \resultlabel{MAE} & \resultlabel{Across models} & \resultci{$+$6.25}{$+$6.07}{$+$6.44} & \resultci{$+$4.01}{$+$3.89}{$+$4.14} \\[2pt]
\resultlabel{Cost} & \resultlabel{MSE} & \resultlabel{None} & \resultci{$+$59.01}{$+$58.11}{$+$59.95} & \resultci{\tableminus 2.61}{\tableminus 3.21}{\tableminus 2.03} \\[2pt]
\resultlabel{Cost} & \resultlabel{MAE} & \resultlabel{None} & \resultci{$+$44.67}{$+$44.19}{$+$45.21} & \resultci{\tableminus 4.11}{\tableminus 4.42}{\tableminus 3.78} \\[2pt]
\resultlabel{Cost} & \resultlabel{MSE} & \resultlabel{Across models} & \resultci{$+$53.16}{$+$52.15}{$+$54.21} & \resultci{\tableminus 3.21}{\tableminus 3.83}{\tableminus 2.63} \\[2pt]
\resultlabel{Cost} & \resultlabel{MAE} & \resultlabel{Across models} & \resultci{$+$41.35}{$+$40.80}{$+$41.94} & \resultci{\tableminus 3.60}{\tableminus 3.92}{\tableminus 3.28} \\[2pt]
\bottomrule
\end{tabular}
\end{table}

\subsection{Answer-format control}\label{app:answer-format-control}
For each test query, we restrict the training pool to queries with the same coarse answer-format label, then retrieve its 100 nearest GTE neighbors.
The two labels, multiple-choice and free-form, are defined in Appendix~\ref{app:answer-format}.
The query's entire family remains excluded, and every restricted training pool contains at least 100 queries.
We use the known task-format label, without generating or classifying any answers.
This control changes both format agreement and the available source-family composition; it does not isolate answer format as the sole cause.

The answer-format restriction raises GTE utility to 0.6974 on RouterBench and 0.7584 on SPROUT, closing approximately 92\% and 27\% of the observed GTE--L21 gaps.
Both restricted-GTE means remain below L21, with paired difference intervals excluding zero.
These results support a substantial role for the format restriction on RouterBench and additional useful reward information in L21 on SPROUT.

Tables~\ref{tab:cost-reward-mean-ci} and~\ref{tab:cost-reward-differences} report these means and paired differences; Tables~\ref{tab:cost-reward-routerbench} and~\ref{tab:cost-reward-sprout} give the results at every penalty.

\subsection{Combining reward and cost estimates}\label{app:combined-estimates}
We evaluate all four combinations of GTE and L21 reward and cost predictions.
At each penalty $\lambda$, every combination selects $\arg\max_m[\widehat r(x,m)-\lambda\widehat c(x,m)]$.
The GTE answer-format control uses both estimates from its restricted neighborhood.
We average recorded routing utility equally over test queries and then equally over the thirteen penalties.
Tables~\ref{tab:cost-reward-mean-ci} and~\ref{tab:cost-reward-differences} give means and paired differences; Tables~\ref{tab:cost-reward-routerbench} and~\ref{tab:cost-reward-sprout} retain all penalties.

\begin{table}[H]
\setlength{\abovecaptionskip}{4pt}
\setlength{\belowcaptionskip}{6pt}
\centering\small
\setlength{\tabcolsep}{5pt}
\caption{Mean out-of-distribution utility over thirteen penalties, with 95\% intervals.}
\label{tab:cost-reward-mean-ci}
\begin{tabular}{@{}llrr@{}}
\toprule
Reward estimate & Cost estimate & RouterBench & SPROUT \\
\midrule
\resultlabel{GTE} & \resultlabel{GTE} & \resultci{0.6615}{0.6583}{0.6651} & \resultci{0.7529}{0.7498}{0.7559} \\[2pt]
\resultlabel{L21} & \resultlabel{GTE} & \resultci{0.6754}{0.6720}{0.6791} & \resultci{0.7709}{0.7679}{0.7739} \\[2pt]
\resultlabel{GTE} & \resultlabel{L21} & \resultci{0.6965}{0.6934}{0.7000} & \resultci{0.7560}{0.7529}{0.7590} \\[2pt]
\resultlabel{L21} & \resultlabel{L21} & \resultci{0.7007}{0.6972}{0.7043} & \resultci{0.7731}{0.7701}{0.7760} \\[2pt]
\multicolumn{2}{l}{\resultlabel{GTE with matching answer format}} & \resultci{0.6974}{0.6942}{0.7010} & \resultci{0.7584}{0.7553}{0.7614} \\[2pt]
\multicolumn{2}{l}{\resultlabel{Training-model means}} & \resultci{0.6618}{0.6586}{0.6652} & \resultci{0.7575}{0.7544}{0.7604} \\[2pt]
\bottomrule
\end{tabular}
\end{table}

\begin{table}[H]
\setlength{\abovecaptionskip}{4pt}
\setlength{\belowcaptionskip}{6pt}
\centering\small
\setlength{\tabcolsep}{5pt}
\caption{Mean utility differences, with paired 95\% intervals. GTE and L21 alone denote ordinary kNN-100.}
\label{tab:cost-reward-differences}
\begin{tabular}{@{}lrr@{}}
\toprule
Comparison & RouterBench & SPROUT \\
\midrule
\resultlabel{Replace GTE reward estimate with L21} & \resultci{$+$0.0139}{$+$0.0120}{$+$0.0156} & \resultci{$+$0.0180}{$+$0.0163}{$+$0.0198} \\[2pt]
\resultlabel{Replace GTE cost estimate with L21} & \resultci{$+$0.0350}{$+$0.0342}{$+$0.0357} & \resultci{$+$0.0031}{$+$0.0027}{$+$0.0036} \\[2pt]
\resultlabel{Replace both GTE estimates with L21} & \resultci{$+$0.0391}{$+$0.0371}{$+$0.0410} & \resultci{$+$0.0202}{$+$0.0185}{$+$0.0220} \\[2pt]
\resultlabel{GTE with matching format minus GTE} & \resultci{$+$0.0359}{$+$0.0350}{$+$0.0368} & \resultci{$+$0.0055}{$+$0.0042}{$+$0.0069} \\[2pt]
\resultlabel{GTE with matching format minus L21} & \resultci{\tableminus 0.0032}{\tableminus 0.0050}{\tableminus 0.0015} & \resultci{\tableminus 0.0147}{\tableminus 0.0163}{\tableminus 0.0131} \\[2pt]
\bottomrule
\end{tabular}
\end{table}

\paragraph{Results.}
On RouterBench, replacing only GTE's cost predictions raises mean utility from 0.6615 to 0.6965, whereas replacing only reward predictions gives 0.6754.
On SPROUT, replacing reward predictions raises utility from 0.7529 to 0.7709, whereas replacing only cost predictions gives 0.7560.
Thus, the contributions differ across benchmarks and are not additive.
SPROUT's L21 cost estimates have slightly higher MSE and MAE, yet still produce a small routing gain, illustrating that aggregate prediction error does not determine routing utility.

\begin{table}[H]
\setlength{\abovecaptionskip}{4pt}
\setlength{\belowcaptionskip}{6pt}
\centering\small
\setlength{\tabcolsep}{5pt}
\caption{RouterBench: utility at every penalty, with 95\% intervals. The first four columns identify the reward/cost predictors; the last uses GTE neighbors matching the query answer format.}
\label{tab:cost-reward-routerbench}
\begin{tabular}{@{}rrrrrr@{}}
\toprule
$\lambda$ & GTE/GTE & L21/GTE & GTE/L21 & L21/L21 & GTE, matching format \\
\midrule
\resultlabel{0} & \resultci{0.7900}{0.7866}{0.7937} & \resultci{0.7854}{0.7818}{0.7891} & \resultci{0.7900}{0.7866}{0.7937} & \resultci{0.7854}{0.7818}{0.7891} & \resultci{0.7902}{0.7868}{0.7939} \\[2pt]
\resultlabel{1} & \resultci{0.7864}{0.7831}{0.7902} & \resultci{0.7804}{0.7768}{0.7842} & \resultci{0.7865}{0.7832}{0.7903} & \resultci{0.7805}{0.7769}{0.7843} & \resultci{0.7867}{0.7834}{0.7905} \\[2pt]
\resultlabel{2} & \resultci{0.7831}{0.7798}{0.7869} & \resultci{0.7774}{0.7738}{0.7812} & \resultci{0.7832}{0.7799}{0.7870} & \resultci{0.7771}{0.7734}{0.7809} & \resultci{0.7832}{0.7799}{0.7870} \\[2pt]
\resultlabel{5} & \resultci{0.7732}{0.7699}{0.7770} & \resultci{0.7676}{0.7640}{0.7713} & \resultci{0.7732}{0.7698}{0.7770} & \resultci{0.7678}{0.7642}{0.7716} & \resultci{0.7726}{0.7690}{0.7764} \\[2pt]
\resultlabel{10} & \resultci{0.7563}{0.7529}{0.7601} & \resultci{0.7498}{0.7462}{0.7536} & \resultci{0.7557}{0.7524}{0.7595} & \resultci{0.7587}{0.7549}{0.7626} & \resultci{0.7638}{0.7601}{0.7676} \\[2pt]
\resultlabel{20} & \resultci{0.7225}{0.7192}{0.7263} & \resultci{0.7180}{0.7144}{0.7219} & \resultci{0.7392}{0.7358}{0.7430} & \resultci{0.7439}{0.7401}{0.7478} & \resultci{0.7491}{0.7456}{0.7529} \\[2pt]
\resultlabel{30} & \resultci{0.6881}{0.6846}{0.6919} & \resultci{0.6859}{0.6823}{0.6900} & \resultci{0.7247}{0.7213}{0.7286} & \resultci{0.7279}{0.7240}{0.7319} & \resultci{0.7341}{0.7306}{0.7380} \\[2pt]
\resultlabel{50} & \resultci{0.6225}{0.6188}{0.6264} & \resultci{0.6226}{0.6187}{0.6269} & \resultci{0.6948}{0.6912}{0.6987} & \resultci{0.6994}{0.6954}{0.7037} & \resultci{0.7029}{0.6991}{0.7069} \\[2pt]
\resultlabel{75} & \resultci{0.5491}{0.5453}{0.5535} & \resultci{0.5619}{0.5575}{0.5662} & \resultci{0.6574}{0.6537}{0.6616} & \resultci{0.6691}{0.6648}{0.6735} & \resultci{0.6638}{0.6600}{0.6680} \\[2pt]
\resultlabel{100} & \resultci{0.5057}{0.5013}{0.5102} & \resultci{0.5945}{0.5900}{0.5994} & \resultci{0.6253}{0.6215}{0.6296} & \resultci{0.6462}{0.6418}{0.6508} & \resultci{0.6334}{0.6296}{0.6377} \\[2pt]
\resultlabel{150} & \resultci{0.5320}{0.5270}{0.5370} & \resultci{0.6150}{0.6104}{0.6194} & \resultci{0.5967}{0.5922}{0.6013} & \resultci{0.6189}{0.6143}{0.6235} & \resultci{0.5937}{0.5897}{0.5981} \\[2pt]
\resultlabel{250} & \resultci{0.5560}{0.5513}{0.5605} & \resultci{0.5859}{0.5813}{0.5905} & \resultci{0.5843}{0.5797}{0.5891} & \resultci{0.5899}{0.5852}{0.5946} & \resultci{0.5592}{0.5546}{0.5638} \\[2pt]
\resultlabel{500} & \resultci{0.5353}{0.5307}{0.5398} & \resultci{0.5362}{0.5316}{0.5408} & \resultci{0.5434}{0.5388}{0.5481} & \resultci{0.5441}{0.5394}{0.5487} & \resultci{0.5341}{0.5296}{0.5386} \\[2pt]
\midrule
\resultlabel{Mean} & \resultci{0.6615}{0.6583}{0.6651} & \resultci{0.6754}{0.6720}{0.6791} & \resultci{0.6965}{0.6934}{0.7000} & \resultci{0.7007}{0.6972}{0.7043} & \resultci{0.6974}{0.6942}{0.7010} \\[2pt]
\bottomrule
\end{tabular}
\end{table}

\begin{table}[H]
\setlength{\abovecaptionskip}{4pt}
\setlength{\belowcaptionskip}{6pt}
\centering\small
\setlength{\tabcolsep}{5pt}
\caption{SPROUT: utility at every penalty, with 95\% intervals. The first four columns identify the reward/cost predictors; the last uses GTE neighbors matching the query answer format.}
\label{tab:cost-reward-sprout}
\begin{tabular}{@{}rrrrrr@{}}
\toprule
$\lambda$ & GTE/GTE & L21/GTE & GTE/L21 & L21/L21 & GTE, matching format \\
\midrule
\resultlabel{0} & \resultci{0.8035}{0.8002}{0.8067} & \resultci{0.8243}{0.8212}{0.8275} & \resultci{0.8035}{0.8002}{0.8067} & \resultci{0.8243}{0.8212}{0.8275} & \resultci{0.7972}{0.7938}{0.8005} \\[2pt]
\resultlabel{0.5} & \resultci{0.8013}{0.7980}{0.8046} & \resultci{0.8224}{0.8193}{0.8256} & \resultci{0.8018}{0.7986}{0.8051} & \resultci{0.8225}{0.8192}{0.8257} & \resultci{0.7956}{0.7923}{0.7990} \\[2pt]
\resultlabel{1} & \resultci{0.7994}{0.7962}{0.8026} & \resultci{0.8204}{0.8173}{0.8235} & \resultci{0.7997}{0.7964}{0.8030} & \resultci{0.8200}{0.8169}{0.8232} & \resultci{0.7951}{0.7917}{0.7984} \\[2pt]
\resultlabel{2} & \resultci{0.7955}{0.7922}{0.7988} & \resultci{0.8165}{0.8132}{0.8196} & \resultci{0.7962}{0.7929}{0.7995} & \resultci{0.8170}{0.8137}{0.8202} & \resultci{0.7940}{0.7906}{0.7973} \\[2pt]
\resultlabel{5} & \resultci{0.7873}{0.7839}{0.7906} & \resultci{0.8075}{0.8041}{0.8108} & \resultci{0.7877}{0.7844}{0.7910} & \resultci{0.8073}{0.8040}{0.8106} & \resultci{0.7907}{0.7874}{0.7940} \\[2pt]
\resultlabel{10} & \resultci{0.7820}{0.7787}{0.7855} & \resultci{0.8004}{0.7970}{0.8037} & \resultci{0.7820}{0.7786}{0.7853} & \resultci{0.8009}{0.7974}{0.8042} & \resultci{0.7878}{0.7844}{0.7911} \\[2pt]
\resultlabel{25} & \resultci{0.7734}{0.7698}{0.7767} & \resultci{0.7926}{0.7892}{0.7960} & \resultci{0.7742}{0.7708}{0.7775} & \resultci{0.7946}{0.7911}{0.7979} & \resultci{0.7816}{0.7782}{0.7849} \\[2pt]
\resultlabel{50} & \resultci{0.7607}{0.7573}{0.7641} & \resultci{0.7795}{0.7761}{0.7829} & \resultci{0.7631}{0.7597}{0.7665} & \resultci{0.7830}{0.7796}{0.7864} & \resultci{0.7714}{0.7679}{0.7747} \\[2pt]
\resultlabel{75} & \resultci{0.7496}{0.7462}{0.7530} & \resultci{0.7671}{0.7637}{0.7706} & \resultci{0.7535}{0.7500}{0.7568} & \resultci{0.7716}{0.7682}{0.7751} & \resultci{0.7606}{0.7573}{0.7641} \\[2pt]
\resultlabel{100} & \resultci{0.7379}{0.7345}{0.7413} & \resultci{0.7555}{0.7520}{0.7589} & \resultci{0.7431}{0.7396}{0.7465} & \resultci{0.7606}{0.7571}{0.7641} & \resultci{0.7497}{0.7464}{0.7531} \\[2pt]
\resultlabel{150} & \resultci{0.7149}{0.7114}{0.7184} & \resultci{0.7318}{0.7282}{0.7353} & \resultci{0.7233}{0.7198}{0.7268} & \resultci{0.7384}{0.7348}{0.7419} & \resultci{0.7283}{0.7249}{0.7318} \\[2pt]
\resultlabel{250} & \resultci{0.6786}{0.6748}{0.6824} & \resultci{0.6924}{0.6886}{0.6963} & \resultci{0.6929}{0.6891}{0.6966} & \resultci{0.6986}{0.6947}{0.7023} & \resultci{0.6904}{0.6865}{0.6942} \\[2pt]
\resultlabel{500} & \resultci{0.6037}{0.5996}{0.6080} & \resultci{0.6115}{0.6071}{0.6157} & \resultci{0.6074}{0.6033}{0.6115} & \resultci{0.6119}{0.6079}{0.6160} & \resultci{0.6165}{0.6123}{0.6206} \\[2pt]
\midrule
\resultlabel{Mean} & \resultci{0.7529}{0.7498}{0.7559} & \resultci{0.7709}{0.7679}{0.7739} & \resultci{0.7560}{0.7529}{0.7590} & \resultci{0.7731}{0.7701}{0.7760} & \resultci{0.7584}{0.7553}{0.7614} \\[2pt]
\bottomrule
\end{tabular}
\end{table}

\end{document}